\documentclass[journal]{IEEEtran}

\usepackage{xcolor,soul,framed} 
\usepackage{multirow,bigstrut,booktabs}
\usepackage{booktabs ,float}
\colorlet{shadecolor}{yellow}
\usepackage{stfloats}
\usepackage{subfig}
\usepackage[pdftex]{graphicx}
\graphicspath{{../pdf/}{../jpeg/}}
\DeclareGraphicsExtensions{.pdf,.jpeg,.png}

\usepackage[cmex10]{amsmath}
\usepackage{array}
\usepackage{mdwmath}
\usepackage{mdwtab}
\usepackage{eqparbox}
\usepackage{url}
\usepackage{algorithmic}
\usepackage{algorithm}
\usepackage{hyperref}
\begin{document}
\bstctlcite{IEEEexample:BSTcontrol}
    \title{Trainable Loss Weights in Super-Resolution}
  \author{{Arash~Chaichi Mellatshahi,
      Shohreh~Kasaei}\\
      {Department of Computer Engineering, Sharif University of Technology }}
      


\maketitle

\begin{abstract}
In recent years, limited research has discussed the loss function in the super-resolution process. The majority of those studies have only used perceptual similarity conventionally. This is while the development of appropriate loss can improve the quality of other methods as well. In this article, a new weighting method for pixel-wise loss is proposed. With the help of this method, it is possible to use trainable weights based on the general structure of the image and its perceptual features while maintaining the advantages of pixel-wise loss. Also, a criterion for comparing weights of loss is introduced so that the weights can be estimated directly by a convolutional neural network. In addition, in this article, the expectation-maximization method is used for the simultaneous estimation super-resolution network and weighting network. In addition, a new activation function, called “FixedSum”, is introduced which can keep the sum of all components of vector constants while keeping the output components between zero and one. As experimental results shows, weighted loss by the proposed method leads to better results than the unweighted loss and weighted loss based on uncertainty in both signal-to-noise and perceptual similarity senses on the state-of-the-art networks. Code is available online\footnote{\href{https://github.com/arashfree/TLW-in-Super-Resolution}{https://github.com/arashfree/TLW-in-Super-Resolution}}.
\end{abstract}

\begin{IEEEkeywords}
super-resolution, loss function, perceptual similarity, expectation-maximization, convolutional neural network. 
\end{IEEEkeywords}

%
\IEEEpeerreviewmaketitle


\section{Introduction}
\IEEEPARstart{T}{oday}, with the increasing application of image super-resolution in enhancing the accuracy and reducing noise in medical images, as well as its growing utilization in improving the quality of computer games, and furthermore, the use of image super-resolution in the analysis and pre-processing of satellite images, the need for improving image quality through deep learning has emerged.
\cite{realworldreview}. 
Fortunately, with the growth of the processing power of GPUs, the possibility of training deep neural networks has become easier and more accessible. As a result, various areas of computer vision and image processing, including the super-resolution problem, have moved towards these networks. The super-resolution problem means finding a high-quality image from a low-quality image. Super-resolution problem is an ill-posed problem with more than one solution for a given input image \cite{survey}. \\
A significant challenge in image super-resolution is the gap between numerical criteria and human judgment. Numerical criteria consider the similarity between the two images due to pixel-wise similarity. But the different pixel of the image does not necessarily have the same importance for human perception \cite{LPIPS}. In this article,  trainable loss weights (TLW) is introduced by estimating the weight for each pixel. Proposed loss uses the advantages of pixel-wise loss and improves the perceptual similarity of images, by weighting the loss in each pixel. For this purpose, a weighting neural network has been needed to estimate the weight of loss. The output of this network should be the fixed sum weights so that each one has a value between zero and one. For this reason, a fixed sum activation layer has been presented. Also, the Expectation-Maximization (EM) \cite{EM} approach has been used to optimize the super-resolution(SR) network and the weighing network. In short, in this paper:
\begin{itemize}
    \item A new method for training the weight of loss has been introduced that improves the learned perceptual image patch similarity (LPIPS) \cite{LPIPS} and signal-to-noise ratio (PSNR) criteria.
    \item A criterion for comparing weights of loss based on the LPIPS network has been defined.
    \item A new activate function, called “FixedSum”, has been introduced which can keep all the components of the outputs between zero and one while keeping the sum of the output array constant.
    \item Optimal weight of loss and high-quality image have been estimated with the EM approach.
\end{itemize}

\begin{figure*}[]
 \begin{center}
\includegraphics[width=0.8\textwidth]{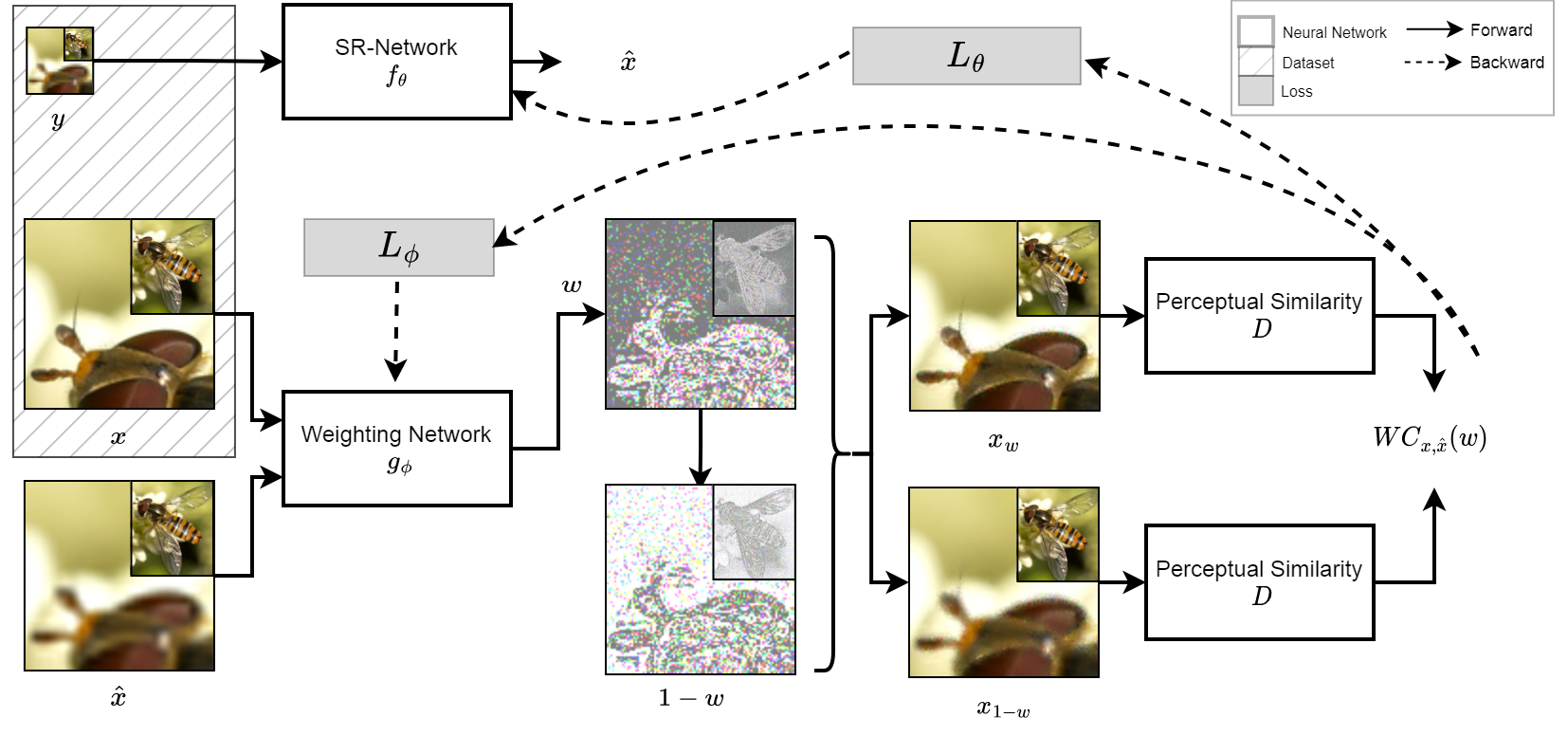}
 \end{center}
\caption{
Proposed training process of SR network and weighting network. After estimating the HR image by the SR network, the weights have been estimated by the weighting network. Then, $x_w$ and $x_{1-w}$ are calculated, and also perceptual similarity network of these images and reference images is estimated. Like this, the weight criterion has been obtained using Eq. (\ref{WeightCriterion}). Next, SR network is trained based on $L_\theta$ which is calculated with the help of Eq. (\ref{sr_loss}). Finally weighting network is trained using $L_\phi$. This loss is obtained based on the negative log Eq. (\ref{optimum_weighting}).}\label{fig:algorithm}
\end{figure*}

\section{Related work}
The results of the super-resolution problem have improved significantly in recent years with the expansion of deep learning networks, and studies in this field have continued with high speed in different sectors. Some studies have focused on the architecture of deep networks and improved various aspects of deep networks, such as sampling and scaling methods \cite{subpixel}\cite{metaSR}, skip connection in very deep network \cite{VDSR}, Attention layer \cite{RCAN}, and Transformers\cite{HAT}. \\
Some other studies have tried to bridge the gap between super-resolution results on training data and real-low-quality images. Some of these studies have considered the method of making low-quality primary images to be the cause of this gap. Usually, a low-quality image has been created by applying a specific degradation method on a high-quality image. As a result, the network that is trained on these pairs has been biased towards the degradation method and cannot work on real images that are degraded differently. 
Non-Blind Super-resolution networks consider the degradation of the image as an input and estimate the high-quality image based on the degradation and the low-quality image \cite{unfolding}\cite{denoiserprior}\cite{multipledegrad}.\\
In non-blind networks, the network is trained with different degradation. So it can meet needs for images with various degradation, but these networks need to have the degradation kernel and noise level, which are unavailable in real images. For this problem, blind super-resolution networks have been introduced, which can simultaneously estimate the degradation and high-resolution image \cite{kernelGAN}\cite{IKC}. Another branch of studies is unsupervised training of networks on unpaired high and low-resolution images \cite{SRGAN}\cite{cincGAN}\cite{Esrgan}. To solve the gap between the results on real images and synthetic degraded images, generative adversarial neural networks have been trained in an unsupervised with unpaired datasets \cite{cincGAN}. In this way, to optimize a super-resolution network, it is enough to take synthetic degraded images to a generative model to transform them into the distribution of real images and train the super-resolution network on the obtained images.Studies on non-blind, blind, and unsupervised super-resolution methods all consider the gap between synthetic and real images to be the root cause of the unfavorable results on general images. They attempt to resolve this issue using unpaired images, estimation of degradation, or degradation input. However, another group of studies, which includes our approach, focuses on developing a new loss function for training networks so that the network can provide a better estimate of super-resolution images based on the structure of the image \cite{LPIPS}\cite{GRAM}\cite{Investigat}\cite{Uncertainty}\cite{Srobb}. \\
Zhang et al \cite{LPIPS} have shown the significant gap between PSNR and structural similarity (SSIM) with human vision and visual perception, then challenge the comparison of models using PSNR and SSIM criteria. Mustafa et al \cite{histogram} have used Earth Mover’s Distance between the RGB histogram of the estimated image and the reference image as an additional loss. They show that combining this loss and regular pixel-wise loss archives better results. Rad et al \cite{Srobb} have discussed perceptual similarity based on the depth of features selected from the perceptual similarity criterion network. They have explained that if the features of the primary layers of the perceptual similarity network have been used as extraction features, then these features are low-level features. As a result, it recognizes images as similar when their edges and margins are more similar. So, if features have been extracted from the middle layers of the perceptual similarity network, the texture images, are compared, and if the features of the last layers of the perceptual similarity network have been extracted, the high-level features of the image, including objects and segments, are compared. Based on this, this article has separated the different parts of the image with the help of a segmentation network by dividing the separate parts of the image into the background, edge, and objects, then it applies an appropriate loss on each section.Abrahamyan et al \cite{GVloss} have employed the gradient of variance of estimated and target images as a new loss. Ning et al \cite{Uncertainty} and Lee et al \cite{GRAM} have used a Bayesian approach to deal with the issue of image super-resolution. Lee et al \cite{GRAM}, have used normalized estimated uncertainty as the weight for maximum likelihood loss. Ning et al \cite{Uncertainty} have taken the uncertainty of each output pixel as a hidden variable in the Bayesian model. Also, by using Jeffrey's prior distribution, which is proportional to the inverse of the uncertainty, have calculated posterior. The posterior is shown as 
\begin{equation}
p(x_i,z_i | y_i)= p(x_i|y_i , z_i)p(\theta_i) \propto \frac{1}{ z_i^2}e^{\frac{\|f(y_i) - x_i\|}{z_i}} \label{eq:unc_posterior}
\end{equation}
In this equation, the pair of images $x$ and $y$ represent the high-quality and low-quality images, respectively. $z$ represents the hidden variable or uncertainty, and $f(.)$ denotes the super-resolution network\cite{Uncertainty}.
 In the first step, the article has funded a network that simultaneously estimates the output image and the uncertainty at each point by optimizing the posterior on the deep network. The loss funtion of first step is shown in the form of
\begin{equation}
L_{ESU} = \frac{1}{N}\Sigma_{i=1}^N e^{s_i}\|f(y_i)-x_i\| + 2s_i b \label{eq:unc_step1}
\end{equation}
where the variable $s_i$ is defined as $s_i = ln(z_i)$ and $N$ is equal to the number of pixels of high-resolution image \cite{Uncertainty}. After sufficient training of this network, in the second step, the loss of each pixel has been weighted according to the uncertainty, and the network has been trained on this weighted loss. The weighted  loss function of the second step of training is also shown in the form of the 
\begin{equation}
L_{UDL} = \frac{1}{N}\Sigma_{i=1}^N s_i\|f(y_i)-x_i\| \label{eq:unc_step2}
\end{equation} \cite{Uncertainty}.
 According to this article, pixels with high uncertainty should be given more weight in the super-resolution problem. This method estimates the weights of loss based on the same conventional loss of image super-resolution and estimates the uncertainty of the loss as weights, unlike our suggested model. So, this method cannot use more complex image features to estimate the weights of loss. Based on our proposed method, the weighting network is trained to estimate loss weights, and the weighting network changes in response to changes in the super-resolution model during training. The proposed method is discussed in the following section.

\section{Proposed Method}
The proposed method has been examined in this section. The proposed criterion for comparing loss weights has been presented first, and then the structure and layers of the weighting network, including the FixSam layer, have been investigated. The following has described the super-resolution model and weighting network's equations and optimization process using an EM approach. The method for training the super-solution network based on the weighted loss of L1 and MSE has been discussed at the end of this section.
\subsection{Criterion for weights of loss}
Typically, SR networks have been learned using L1 and MSE losses. These two pixel-wise losses have been frequently used for super-resolution training due to their simple calculation. Because they have determined the pixel-to-pixel distance error between the model's output and the reference image, these losses are unconcerned with the image's higher-level features. But the perceptual similarity metric has been trained based on human judgments. This attempt has been made to create a metric based on the image's features to determine how similar the two images are. Our approach proposes the weighted L1 and MSE loss, therefore the weights have been calculated by the weighting network. Hence, it is necessary to determine the best weights using the model's output and the reference image. To evaluate the weight of loss quality, a criterion is established based on the LPIPS network. This criterion is shown by 
\begin{equation}\label{WeightCriterion}
\begin{cases}
    x_w = (1-w) \odot \hat{x} + w \odot x \\
    x_{1-w} = w \odot \hat{x} + (1-w) \odot x \\
    WC_{x,\hat{x}}(w) = \frac{D(x_{1-w} , x) + \epsilon}{D(x_{w} , x) + \epsilon}
\end{cases} .
\end{equation}
Where $w$ represents a weight, $x$ denotes a reference image, $\hat{x}$ represents the output image generated by the super-resolution (SR) model, and $D$ refers to the judge network or the pre-trained LPIPS network, responsible for calculating the similarity between two input images. To avoid division by zero and prevent the metric from becoming unbounded, the parameter $\epsilon$ is employed.

As depicted in Figure \ref{fig:algorithm}, when the weight $w$ assigns higher values to important pixels, these pixels are selected from the reference image, resulting in the formation of $x_w$. Consequently, $x_w$ becomes closer to $x$, while $x_{1-w}$ moves farther away from $x$, leading to an increase in the weight criterion ($WC)$. Consequently, optimal weights exhibit higher $WC$ values, whereas non-optimal weights yield lower $WC$ values.

The result from the weight criterion ($WC$) can be used to compare different weights of loss and also for training weighting network. The maximum value of this criterion occurs when the weighting network can select the most influential pixels to make the two images, $x_w$ and $x$, closer to each other, while maximizing the dissimilarity between $x$ and $x_{1-w}$. It should be noted that by similarity, we refer not only to pixel-level similarity but also to structural similarity based on the features that the judge network uses to make decisions.

One obvious solution for optimal weights and weighting networks is to assign equal importance to all image pixels, i.e., considering all weights as equal to 1. This way, $x_w$ and $x$ become identical. However, the weight network should be able to distinguish influential pixels from irrelevant ones. Hence, a constraint is imposed on the sum of weights. The application of this constraint is explained in the following section, considering the structure of the weighting network and the new layer called "FixedSum".

\subsection{Weighting Network}
\begin{figure}

\centering
\subfloat[]{\includegraphics[width=0.4\linewidth]{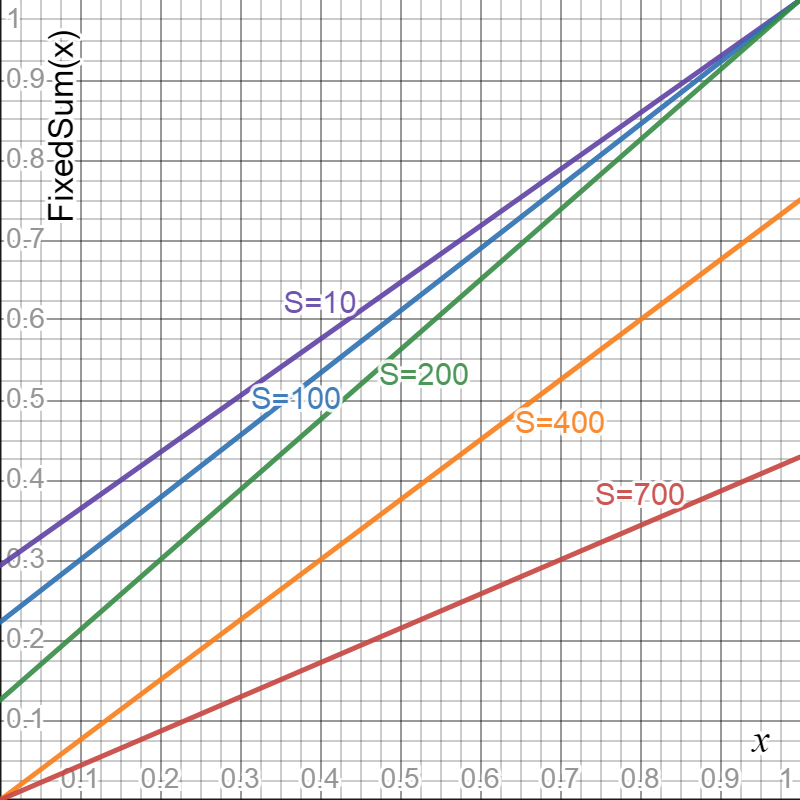} %
\label{fig:fixedsum1}}
\hfil
\subfloat[]{\includegraphics[width=0.4\linewidth]{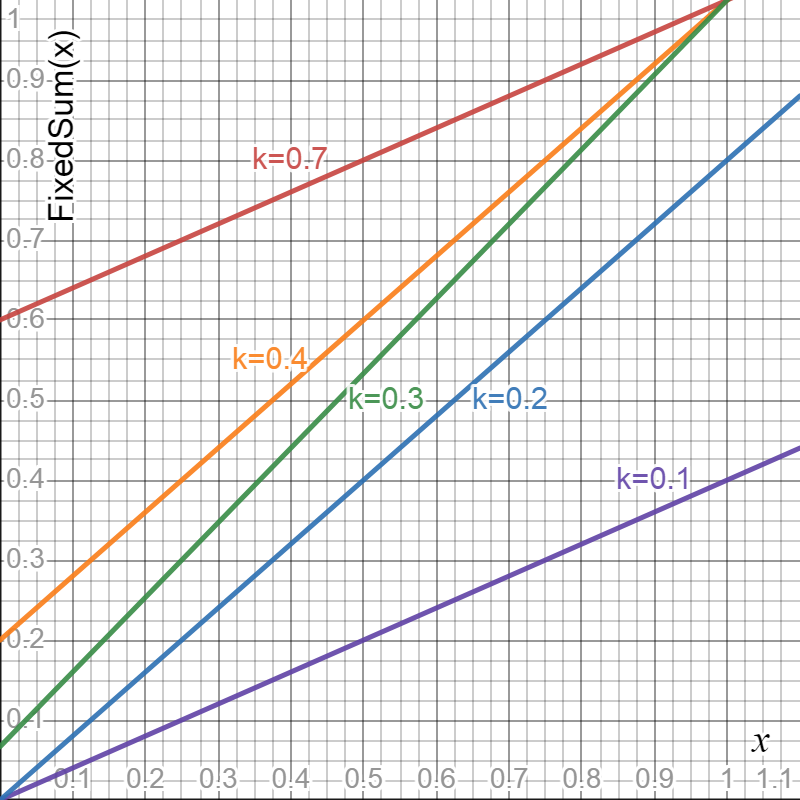}%
\label{fig:fixedsum2}}
\caption{The FixedSum activation function is designed for a scenario where (a) the sum of the weights is constant and equal to a predetermined value, typically denoted as $k=0.3$. In this case, the function is plotted for a given image size of $N=1000$ with different values of $S$. As observed, when the sum of the input values is smaller than $k \cdot N$, the activation function adjusts the output sum to the desired value ($k \cdot N$) using a linear transformation with a bias term applied to the input elements. However, if the input sum is larger, the activation function employs a linear function to adjust the output sum accordingly. (b) when the sum of weights is fixed and equal to a specific value, denoted as $S=250$. In this case, the activation function is plotted for a given image size of $N=1000$ and different values of $k$.}
\label{fixedsum}
 \end{figure}

The weighting network should be able to determine the optimal weight based on the current output of the super-resolution network and the reference image. A four-layer convolutional neural network has been used with the ReLU activation function in the first three layers. The network's output must have a fixed sum, and the values of each pixel must be between 0 and 1. Therefore, the FixedSum activation layer has been introduced. This activation function is represented by 
\begin{equation}\label{fixedsum}
    FixedSum(x,k)  = 
    \begin{cases}
        x + \frac{k\cdot N - S}{N - S} \cdot (1-x) , & k \cdot N > S\\
        x - \frac{S-k\cdot N}{S} \cdot x, & k \cdot N \leq S .
    \end{cases}
\end{equation} 
In this equation, $x$ represents an input vector of the activation function, $S$ represents a sum of the $x$ components, $N$ represents the number of the $x$ vector's components, and $k$ denotes a fixed ratio. Also, The sum of the output vectors is set to $k\cdot N$.As a result, the output obtained from the weighted model has a fixed and smaller sum of weights compared to N, which causes the network to be compelled to differentiate between pixels and identify influential ones. The FixedSum layer is introduced for this purpose, allowing the preservation of a certain sum of output weights while enabling the modification of all elements between zero and one. As demonstrated in Figures \ref{fig:fixedsum1} and \ref{fig:fixedsum2}, if the sum of weights of loss($S$) is smaller than $k\cdot N$, the "FixedSum" activation function attempts to linearly combine the values of each element between zero and one while maintaining the output values, in order to reach a sum of $k\cdot N$. However, if $S$ exceeds the value of $K \cdot N$, the activation function reduces the sum of weights through a linear combination until it reaches the desired value.\\
To examine the effect of the FixedSum activation function on backpropagation, we investigate the derivative of this function with respect to its input. As shown in the equation,

\begin{equation}\label{gradient}
    \small{\frac{\partial FixedSum(x,k)_{i,j}}{\partial x_{i,j}}  = }
    \begin{cases}
        \begin{split}
         \scriptstyle{\frac{N(1-k)(N-S+x_{i,j}-1)}{(N-S)^2}  > 1 - k,} 
        \end{split} &\scriptstyle{k \cdot N > S}   \\

        \begin{split}
              \scriptstyle{\frac{kN(S-x_{i,j})}{S^2}  > k,}
        \end{split} & \scriptstyle{k \cdot N \leq S}
     \end{cases}
\end{equation} 

The derivative of the activation function is greater than $1-k$ when $k \cdot N > S$, and it is greater than $k$ when $k \cdot N \leq S$. Considering that the value of $k$ is determined between 0.3 to 0.9, the derivative will always be greater than 0.1. Therefore, the FixedSum activation function does not lead to network saturation and does not hinder error backpropagation.\\

\begin{figure}
 \includegraphics[width=0.8\linewidth]{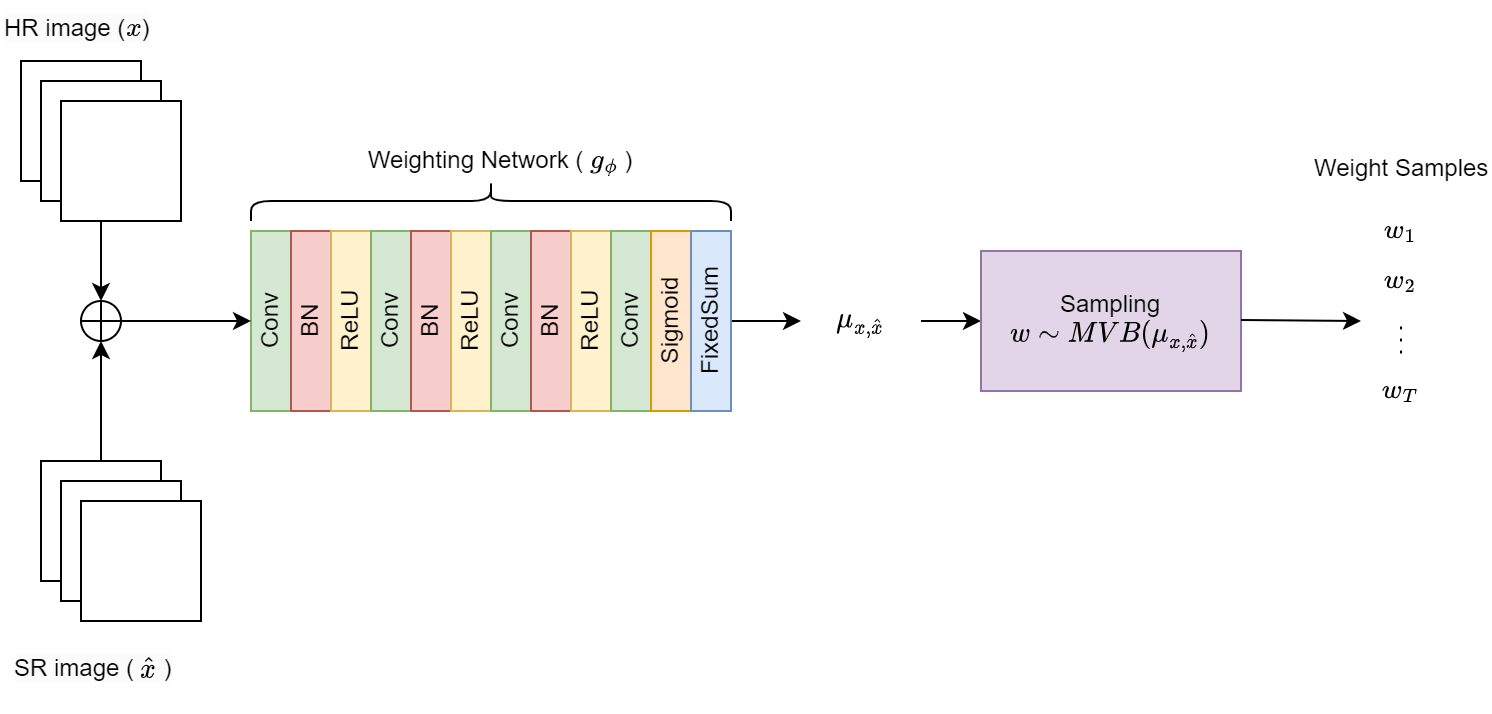} 
\caption {The structure of the weighting network.}
\label{fig:WNet}
\end{figure}

The output of the Weighting Network determines a multivariate Bernoulli distribution (MVB). In this paper, we sample from this distribution to determine the weights of loss. As shown in Figure \ref{fig:WNet}, after calculating the mean of the MVB distribution using the weighting network according to the equation $\mu_{x,\hat{x}}=g_\phi(x,\hat{x})$, each sample from this distribution yields a weight matrix. The multivariate Bernoulli distribution was chosen because the network needs to select influential and non-influential pixels among the image pixels to improve image quality. Therefore, the network should estimate a Bernoulli distribution for each pixel, indicating the probability that the pixel should be selected, based on the current trained SR model. As a result, the weights of loss are obtained by sampling from this distribution after the mean weight distribution, following the equation $w \sim MVB(g_\phi(x,\hat{c}))$. Furthermore, to use the Expectation-Maximization (EM) method in the simultaneous training of the weighting and super-resolution models, it was necessary to sample from the weight distribution. To ensure that the weighting network remains differentiable after sampling, a relaxed Bernoulli distribution \cite{MVB} was used for sampling. The output of the network after the sampling layer remains differentiable and trainable, allowing for backpropagation.

\subsection{EM approach}
The super-resolution problem has been generally defined as finding the maximum likelihood network on the LR-HR data. In this paper, the weight of loss is considered the hidden variable in determining the likelihood of the super-resolution network. Therefore the optimization problem must be solved using the Expectation-maximization method.
In the expectation step, the expected likelihood for the distribution of super-resolution network parameters has been calculated using the previous network and the estimated distribution for the weight of loss. Next, in the maximization step, the most optimal network is selected. The likelihood of super-resolution network parameters ($\theta$) is shown by  
\begin{equation}\label{ikelihood_theta_w}
P(x,y \vert \theta,w)=\frac{1}{z}e^{-\| w \odot (x-f_\theta(y)) \|},
\end{equation}
where $x$ and $y$ are the pair of HR and LR images, $f_\theta(.)$ denotes an SR network, and $w$ denote the weight of loss. If $w$ in this equation is a fixed weight matrix, the maximum likelihood equation becomes equivalent to the conventional weighted loss in a similar form to the equation described in the article \cite{Uncertainty}. Typically, using L1 and MSE errors for training neural networks involves finding the maximum negative log-likelihood on a multivariate normal distribution with the mean of the HR image and a identity covariance matrix. Similarly, it is assumed here that the distribution of SR images follows a multivariate normal distribution with the mean of the HR image and a diagonal covariance matrix, where the diagonal elements correspond to the weight values of $w$. However, since the matrix $w$ itself is a random variable, estimating the maximum likelihood parameters of the SR network requires estimating the distribution of the weight variable $w$ at each stage. In this paper, we define the likelihood of the random variable $w$ based on the estimated $WC$ criterion for it. This distrubution is represented by 
\begin{equation}\label{optimum_weighting}
P(w \vert \theta , x, y)= \frac{1}{z'} WC_{x,f_\theta (y)} (w)
\end{equation}.
To find the optimal weights of loss, it is necessary to estimate the parameters of the $WC$ distribution so that it can provide an appropriate distribution for the weights of loss based on the current state of the SR network and the input image. For this purpose, the likelihood is calculated on the parameters of the weighting network according to to the following equation:
\begin{equation}\label{optimum_phi}
P(\phi \vert \theta , x , y)= \frac{1}{z''} E_{w  \sim MVB(g_\phi (x,f_\theta (y))) } [P(w \vert \theta,x,y)].
\end{equation}.
In these equations, $\phi$ is considered weighting network parameters, $g_{\phi}(.)$ is considered the weighting network, and $g_\phi(x,\hat{x})$ is denoted the mean of this MVB distribution.order to optimize the weighting network, we can achieve it by calculating the negative logarithm of the expected value of the $WC$ variable based on the provided multivariate Bernoulli distribution ($MVB$). This calculated value serves as the loss function $L_\phi$ for training the weighting network. To accomplish this, the loss function is computed by sampling from the aforementioned distribution. in following, using the EM method the expected likelihood of $\theta$ super-resolution network parameters, taking into account the previous network $\theta^{t}$ is represented by 
\begin{equation}\label{expected_step}
Q(\theta \vert \theta^{(t)} )=E_{w \vert \theta^{(t)},x,y} [log P(x,y \vert \theta,w)].
\end{equation} 
in order to determine the loss of the super-resolution model, $T$ weights from the multivariate Bernoulli distribution are re-sampled Then the expected likelihood of a super-resolution network should be approximated by 
\begin{equation}\label{expected_step_estimation}
\begin{split}
&\scriptstyle{E_{w \vert \theta^{(t)},x,y} [log P(x,y \vert \theta,w)]} \\ 
 &\scriptstyle{ \approx \sum_{w  \sim MVB(g_\phi (x,f_\theta (y))) } P(w \vert \theta,x,y) \cdot log P(x,y \vert \theta,w)} \\
&\scriptstyle{ \approx \sum_{w  \sim MVB(g_\phi (x,f_\theta (y))) } - WC_{x,f_\theta (y)} (w)  \cdot {\| w \odot (x-f_\theta(y)) \|}} \\
 &\scriptstyle{\approx - [\sum_{w  \sim MVB(g_\phi (x,f_\theta (y))) } WC_{x,f_\theta (y)} (w) \cdot w ]   \odot {\| x-f_\theta(y) \|}}
\end{split}
\end{equation}.
In the next step, the maximization step of the EM algorithm is denoted by 
\begin{equation}\label{maximization_step}
\theta^{(t+1)}=argmax_\theta Q(\theta \vert \theta^{(t)}).
\end{equation} 
\cite{EM} to optimize the super-resolution network. So, the SR network loss is calculated from 
\begin{equation}\label{sr_loss}
\begin{split}
&L_\theta = argmin_\theta - E_{w \vert \theta^{(t)},x,y}[log P(x,y \vert \theta,w)] \\
&=  [\sum_{w  \sim MVB(g_\phi (x,f_\theta (y))) } WC_{x,f_\theta (y)} (w) \cdot w ]   \odot {\| x-f_\theta(y) \|}
\end{split}
\end{equation}
where $\|.\|$ denotes the basic loss that can indicate L1 loss or MSE loss. In this equation, $WC$ and $w$ are constant numbers and zero derivatives with respect to $\theta$ and are only used as coefficients. The pixel-wise weight $w$ is applied to L1 loss without any change. But applying weight to the MSE loss, considering that this loss actually can be shown as a weighted L1 loss with $\frac{|x-\hat{x}|}{2}$ weight. As a result, applying a new weight that ranges between zero and one will have little effect. So, for better weighting of the MSE loss, the weights of the MSE loss have been modified by $w \gets \frac{w}{|x-\hat{x}| + 0.1} $. Algorithm\ref{alg:one} shows the training process of the super-resolution and weighting model. In each iteration, first, the optimum weighting network has been funded by optimizing the $WC$ based on the previous super-resolution network. For this purpose, according to the output MVB distribution of the weighting model, a weighted sample has been taken and then based on  
Eq. (\ref{optimum_weighting}) the weighted criterion for the sample is calculated. This method is used to estimate $P(\phi \vert \theta , x , y)$ by Eq. (\ref{optimum_phi}). Second, in order to determine the loss of the super-resolution model, $T$ weights from the multivariate Bernoulli distribution are re-sampled. Third,  the expected likelihood of a super-resolution network should be computed by Eq. (\ref{expected_step}). To estimate this value, the mean of $P(w \vert \theta, x, y)P(x,y \vert w,\theta)$ for each sampled weight has been approximated. At the end of the iteration, using the approximated loss the super-resolution network is optimized. 

\begin{algorithm}[H]
\caption{Super-resolution model and weighting model training process.}\label{alg:one}
\begin{algorithmic}

\STATE {\textbf{FOR}} {$x \gets {HR}$ , $y \gets {LR}$}
\STATE \hspace{0.5cm} $\hat{x} \gets f_\theta(y)$\;
\STATE \hspace{0.5cm} \textbf{sample} $w \sim MVB(g_\phi(x,\hat{x}))$\;
\STATE \hspace{0.5cm} $Loss_{\phi} \gets -logWC_{x,\hat{x}}(w)$ \;
\STATE \hspace{0.5cm} \textbf{train} $\phi$ by optimizing $Loss_{\phi}$\;
\STATE \hspace{0.5cm} \textbf{sample} $w^{(1)},w^{(2)},...,w^{(T)}\sim MVB(g_\phi(x,\hat{x}))$\;
\STATE \hspace{0.5cm} $Loss_{\theta} \gets \sum_{i=1}^T WC_{x,\hat{x}}(w^{(i)})  \| w^{(i)} \odot (x-f_\theta(y)) \|$\;
\STATE \hspace{0.5cm} \textbf{train} $\theta$ by optimizing $Loss_{\theta}$\;

 \end{algorithmic}
\end{algorithm}

\section{Experimental Results}

\begin{figure*}[t]    
    \centering

    \begin{tabular}{ccc}

\cline{1-3} 

\tiny{Output of Sigmoid}
& \tiny{Output of Fixedsum $\mu_{x,\hat{x}}$}
& \tiny{Sample Weight $w$} \\

\scalebox{0.18}{\includegraphics{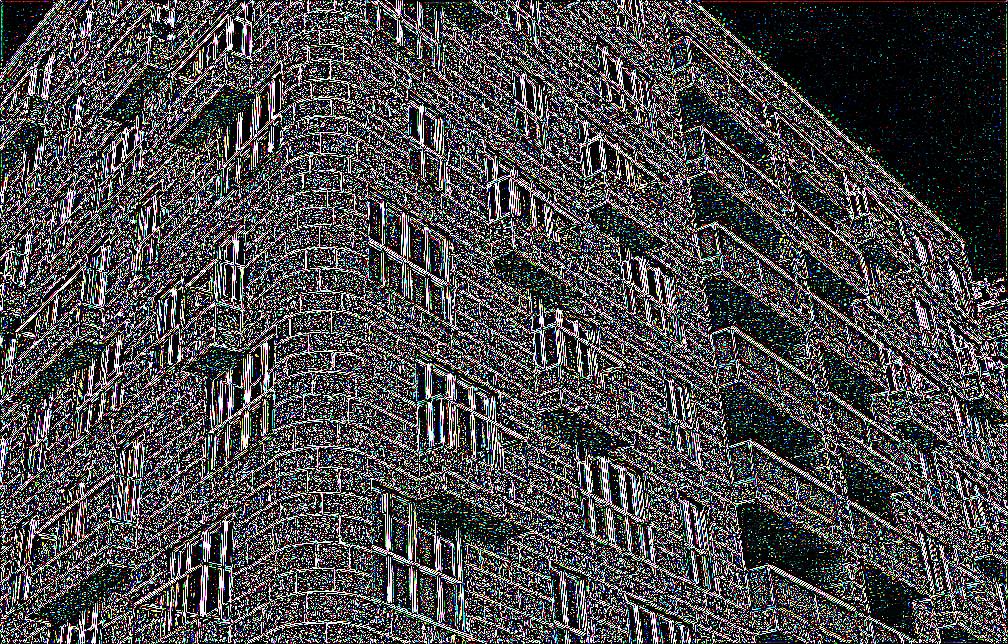}}
& \scalebox{0.18}{\includegraphics{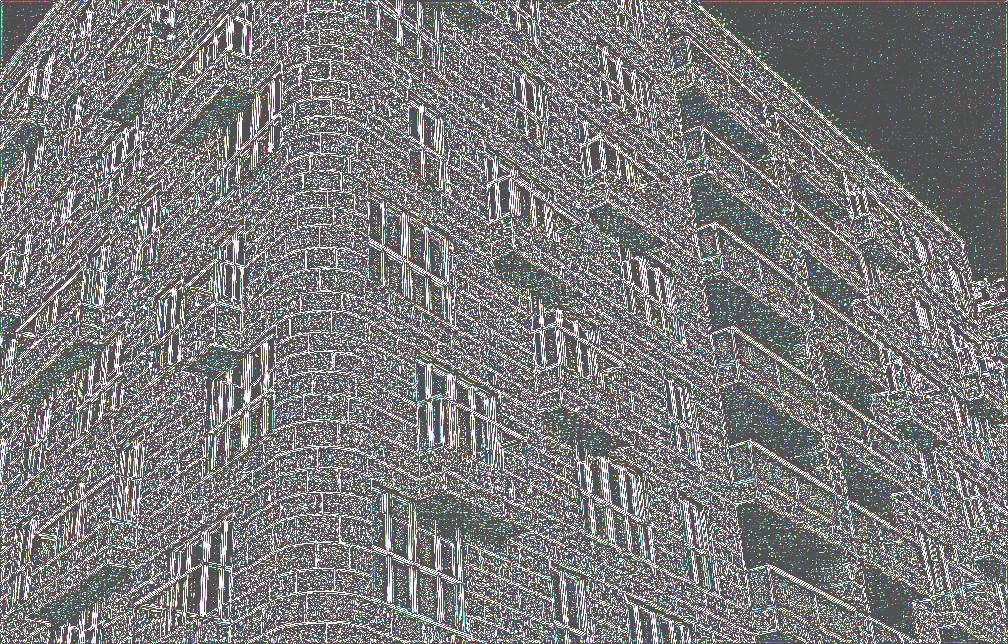}}
& \scalebox{0.18}{\includegraphics{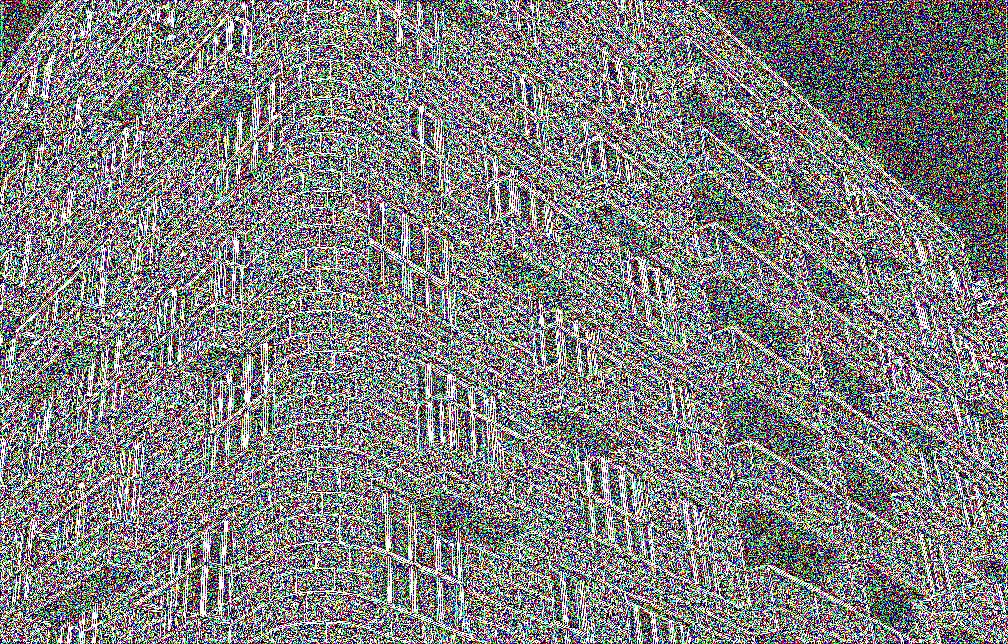}}
\\

\scalebox{0.12}{\includegraphics{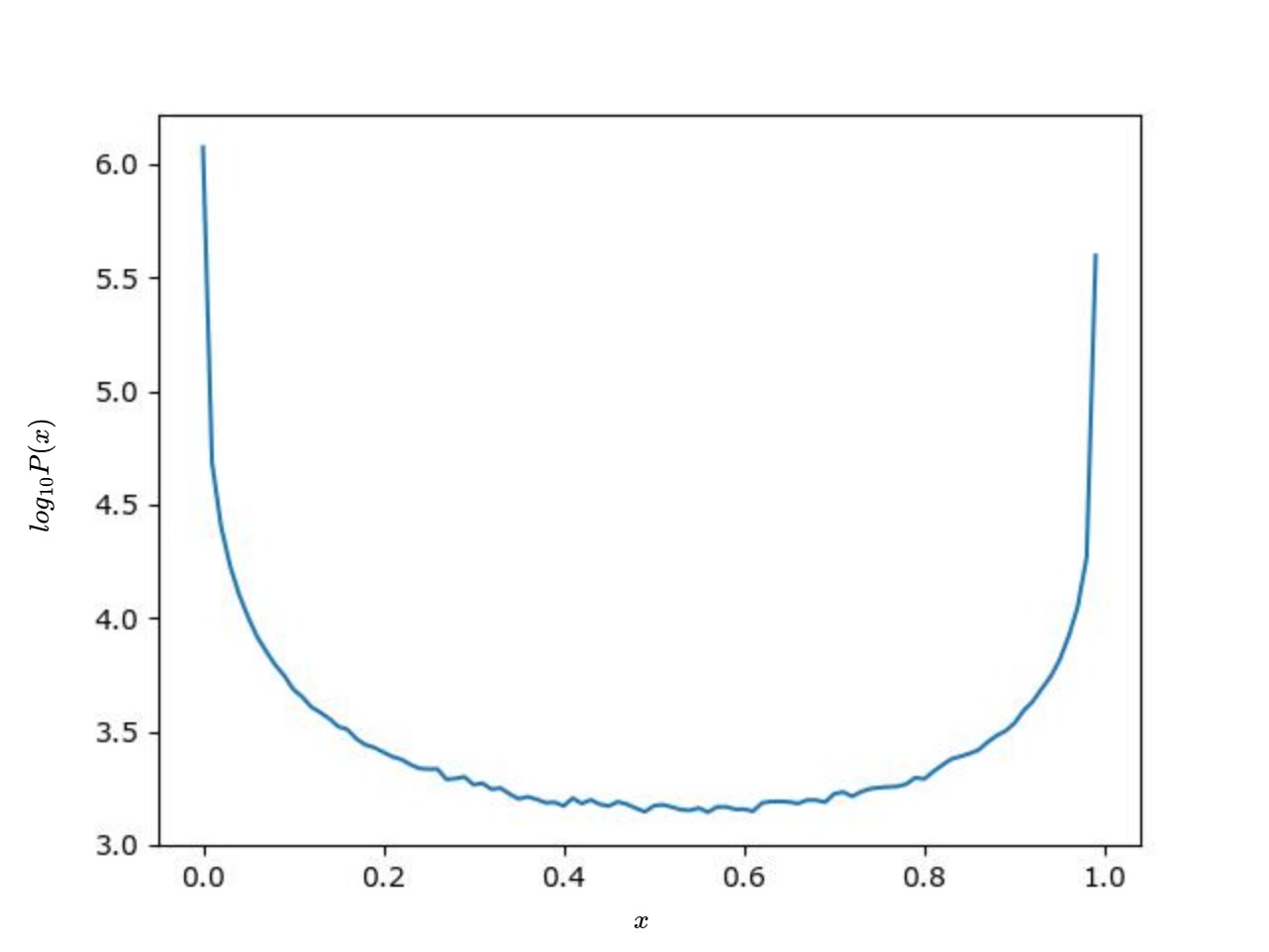}}
& \scalebox{0.12}{\includegraphics{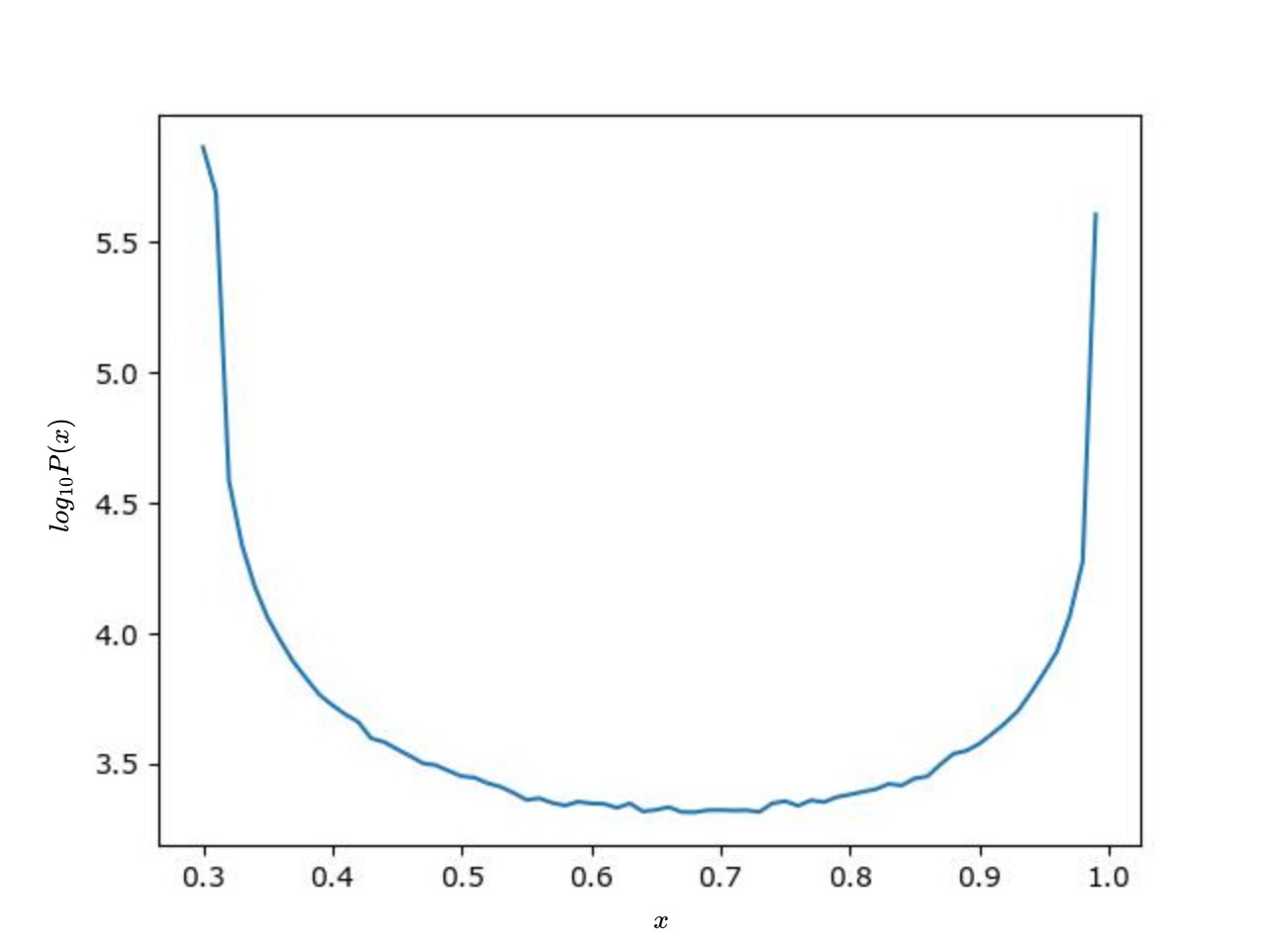}}
& \scalebox{0.12}{\includegraphics{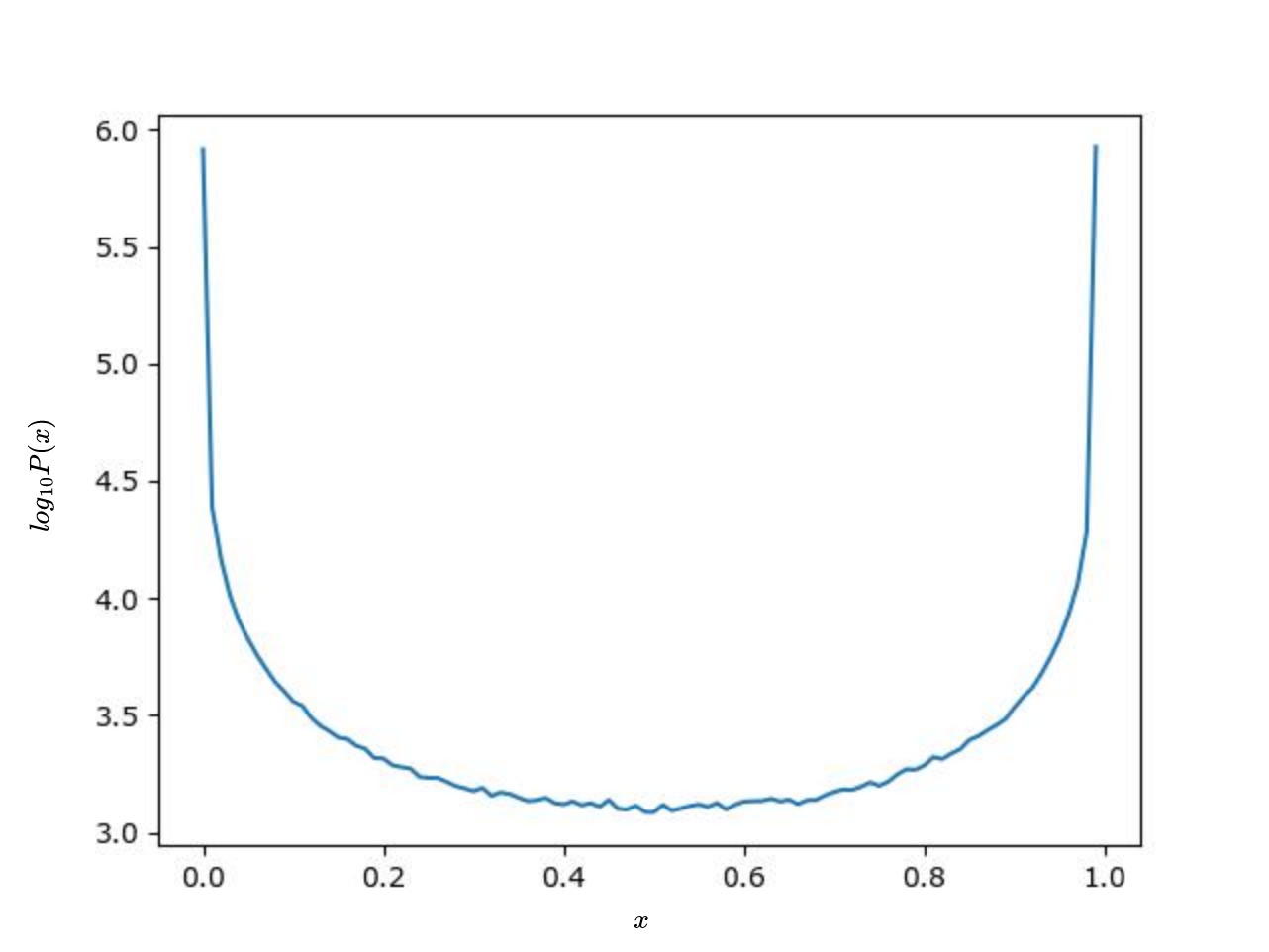}}
\\

   \end{tabular}
   \caption{Examining the Impact of the \textit{FixedSum} Activation Function and Sampling:The first row displays the output of each layer, while the second row illustrates the distribution of output values.}
    \label{fig:wnetvis}
\end{figure*}

\begin{figure*}[]
    \centering
    \includegraphics[width=0.9\textwidth]{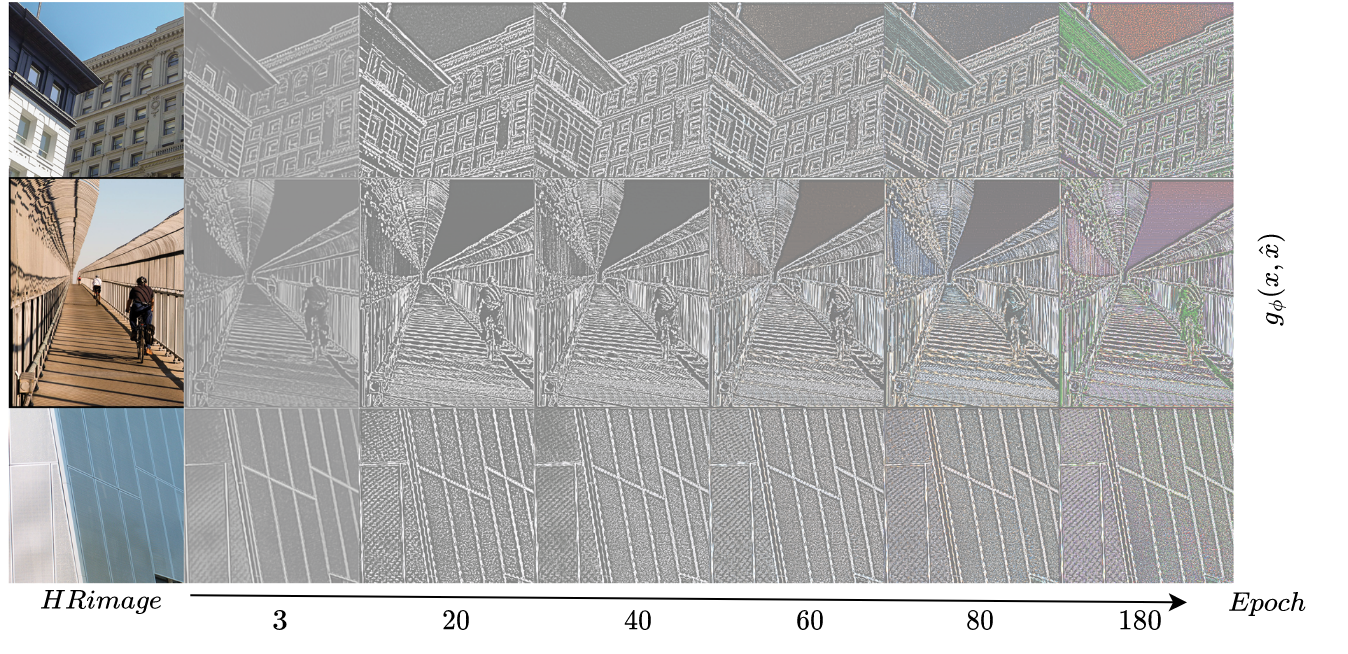}
    \caption{Weight-Epoch diagram; at the beginning of the training process, the network is focused on the edges and low-level features. Subsequently, in addition to the edges, the importance of textures and then colors increases.}
    \label{fig:weight_epoch}
\end{figure*}

\renewcommand{\arraystretch}{1}
\begin{figure*}[t]    
    \centering
    \begin{tabular}{ccccc}

\multicolumn{2}{c}{} 
& 
& 
& \textbf{Proposed method} \\

\cmidrule(lr){5-5}

\multicolumn{2}{c}{\multirow{3}{*}[1.4cm]{\scalebox{.27}{\includegraphics{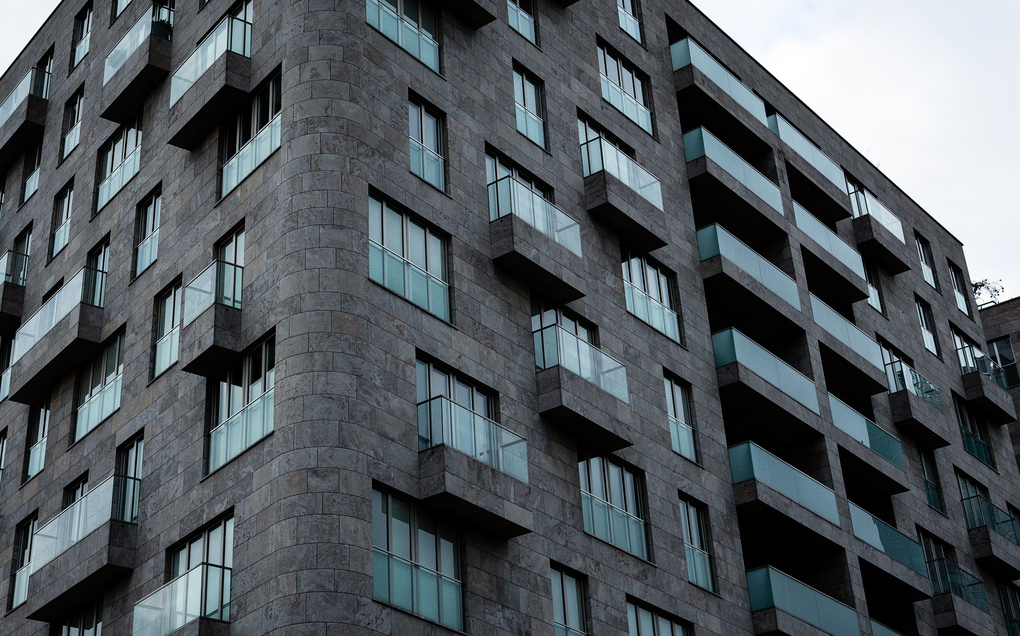}}  }} 
& \scalebox{1.0}{\includegraphics[trim={9cm 9.0cm 13.5cm 5.1cm},clip]{pdf/0hr.png}}
& \scalebox{1.0}{\includegraphics[trim={9cm 9.0cm 13.5cm 5.1cm},clip]{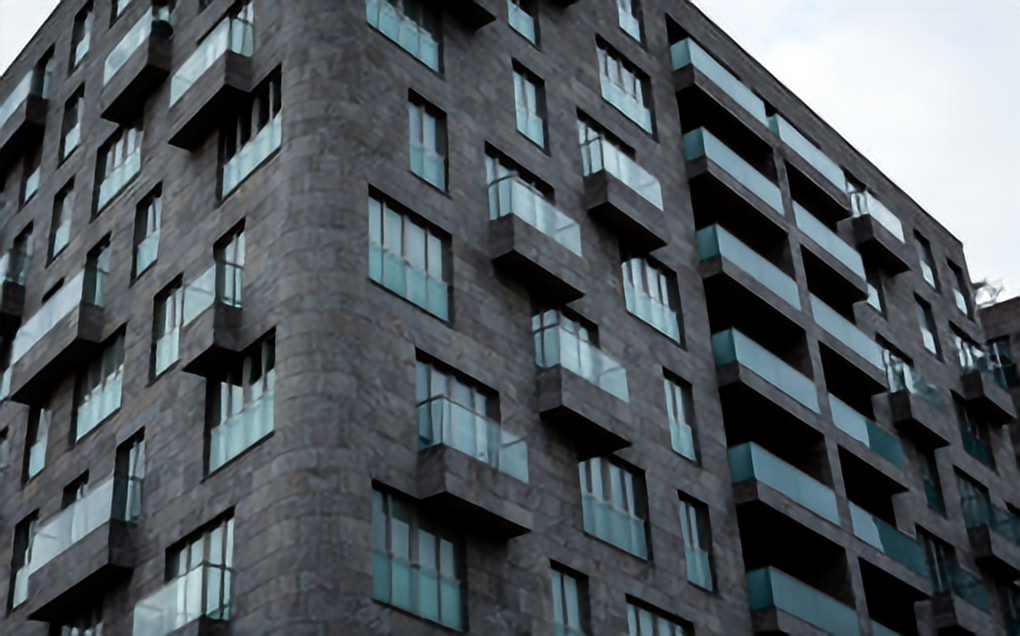}}
& \multicolumn{1}{|c|}{\scalebox{1.0}{\includegraphics[trim={9cm 9.0cm 13.5cm 5.1cm},clip]{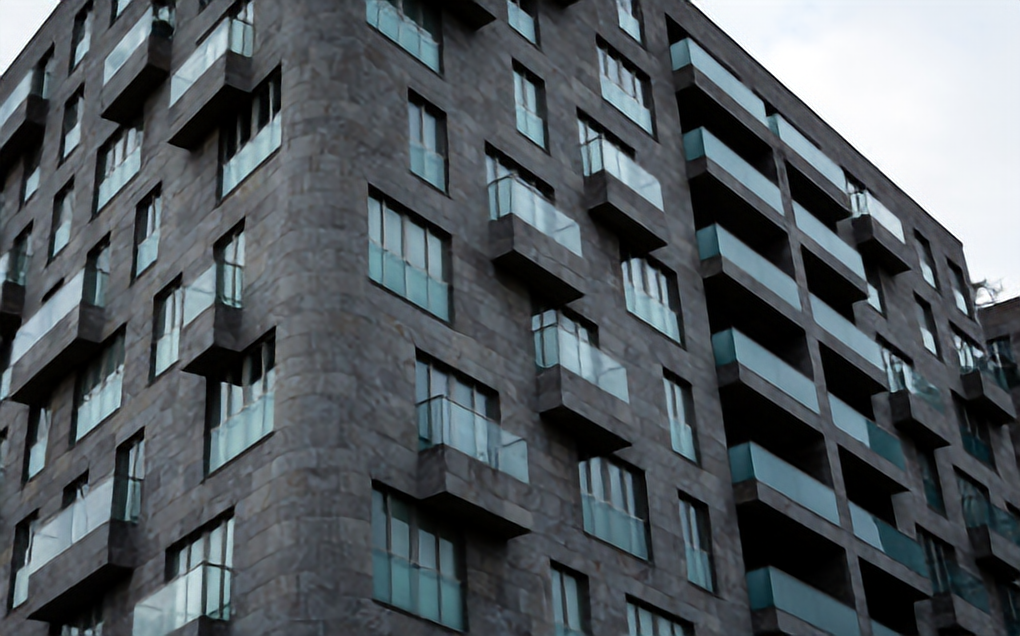}}}
\\
&
& \small{HR} & \small{L1} & \multicolumn{1}{|c|}{\small{TLW+L1}} \\

&
& \scalebox{1.0}{\includegraphics[trim={9cm 9.0cm 13.5cm 5.1cm},clip]{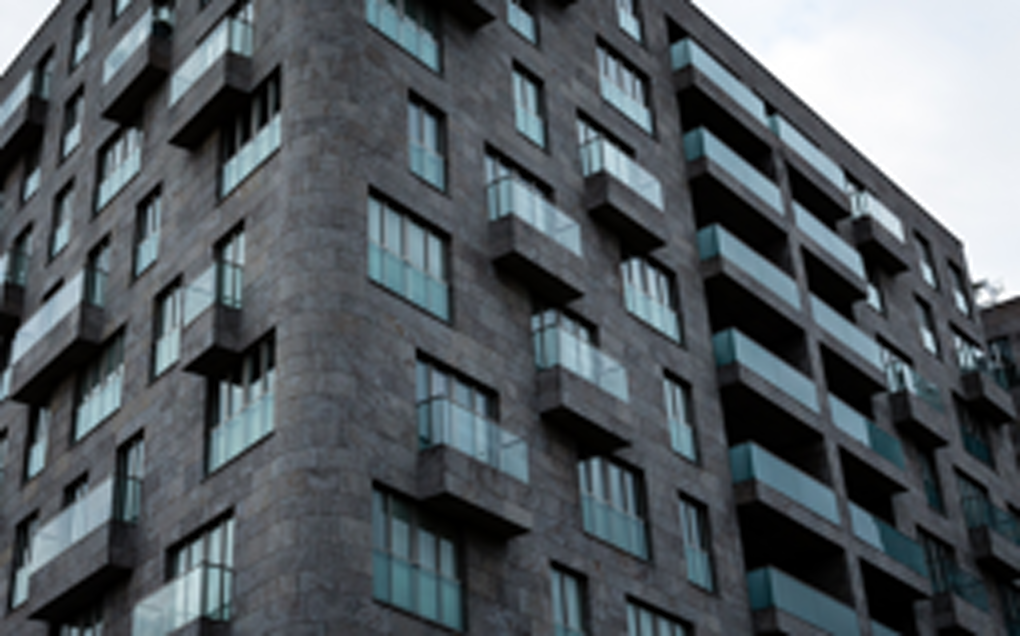}} 
& \scalebox{1.0}{\includegraphics[trim={9cm 9.0cm 13.5cm 5.1cm},clip]{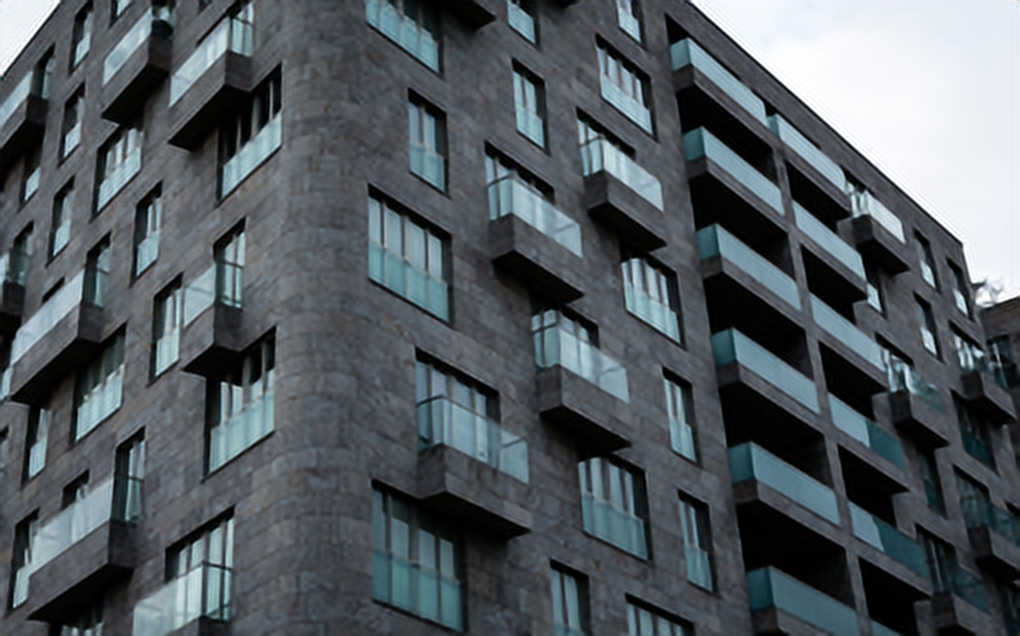}}
& \multicolumn{1}{|c|}{\scalebox{1.0}{\includegraphics[trim={9cm 9.0cm 13.5cm 5.1cm},clip]{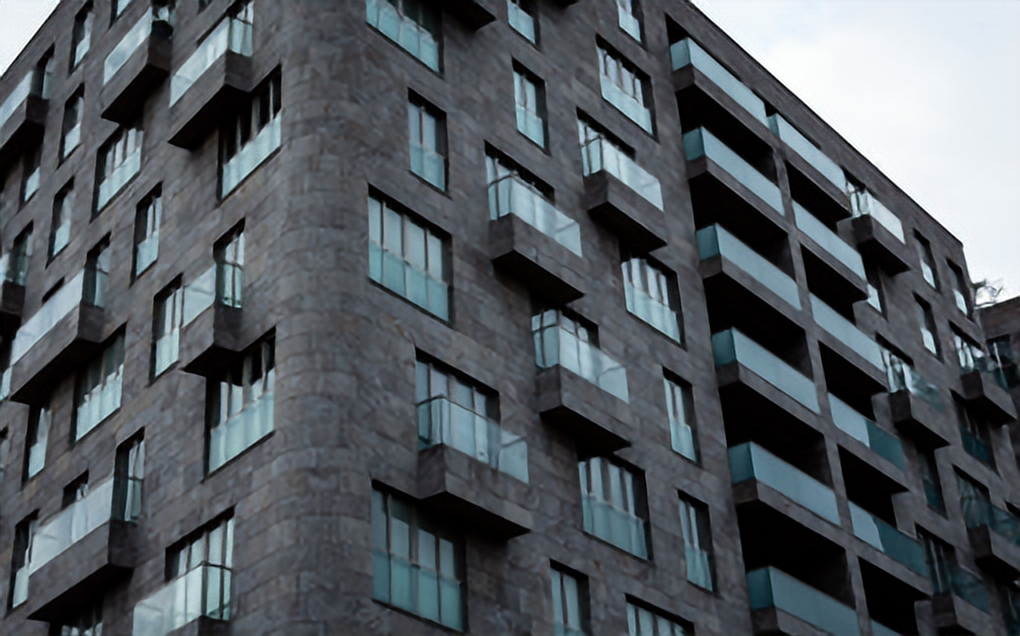}}}
\\
\multicolumn{2}{c}{} 
& \small{BICUBIC} & \small{MSE} & \multicolumn{1}{|c|}{\small{TLW+MSE}} \\

\multicolumn{2}{c}{\multirow{3}{*}[1.6cm]{\scalebox{.27}{\includegraphics{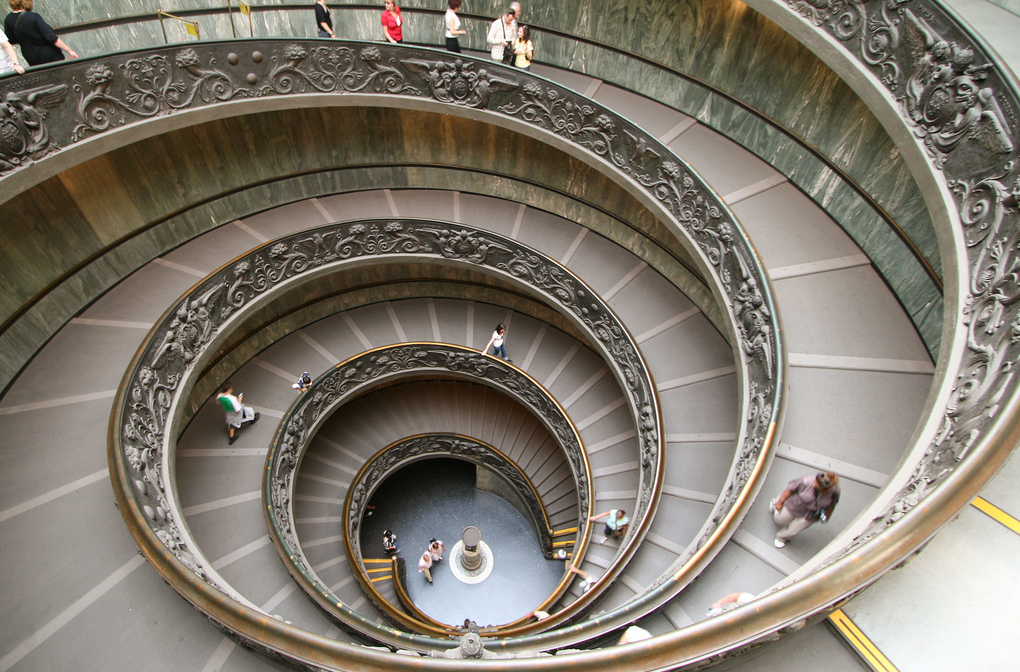}}}} 
& \scalebox{1.0}{\includegraphics[trim={11cm 15.0cm 11.5cm 0cm},clip]{pdf/6hr.png}} 
& \scalebox{1.0}{\includegraphics[trim={11cm 15.0cm 11.5cm 0cm},clip]{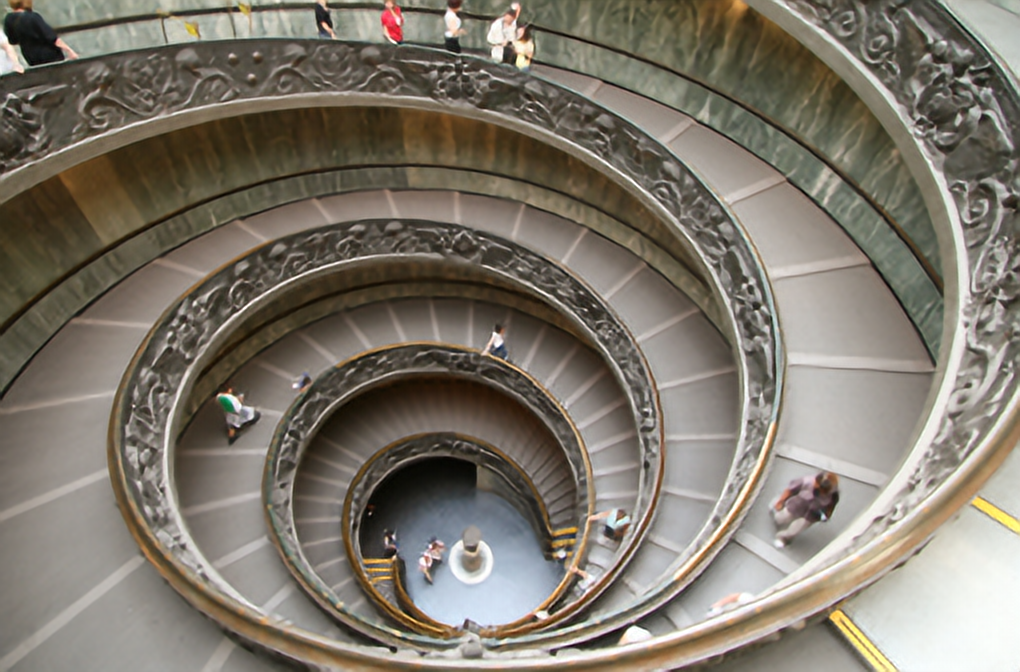}} 
& \multicolumn{1}{|c|}{\scalebox{1.0}{\includegraphics[trim={11cm 15.0cm 11.5cm 0cm},clip]{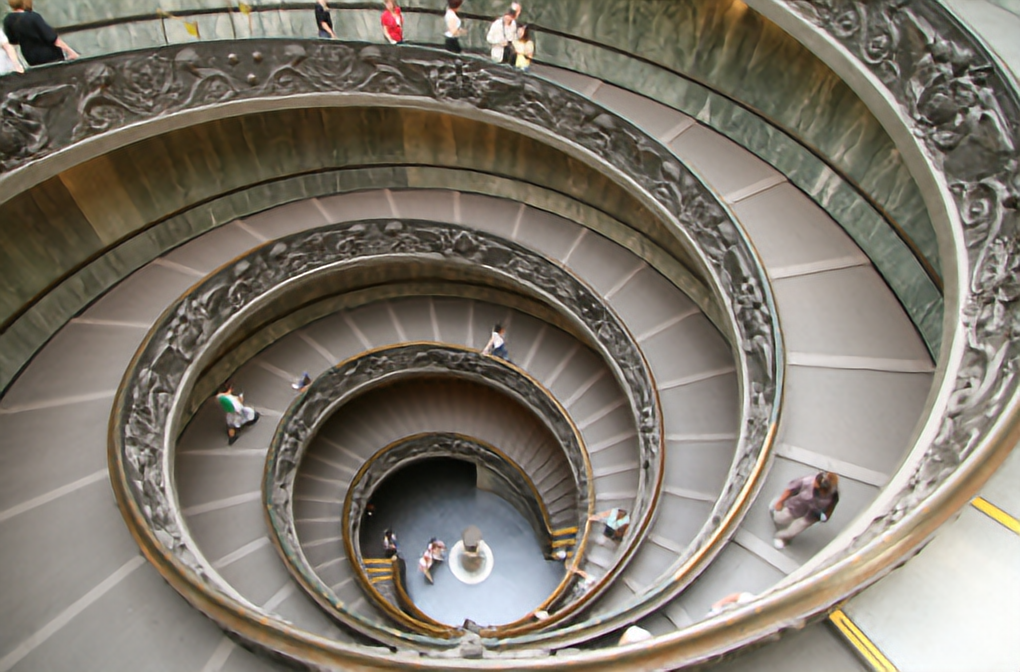}}}
\\
&
& \small{HR} & \small{L1} & \multicolumn{1}{|c|}{\small{TLW+L1}} \\

&
& \scalebox{1.0}{\includegraphics[trim={11cm 15.0cm 11.5cm 0cm},clip]{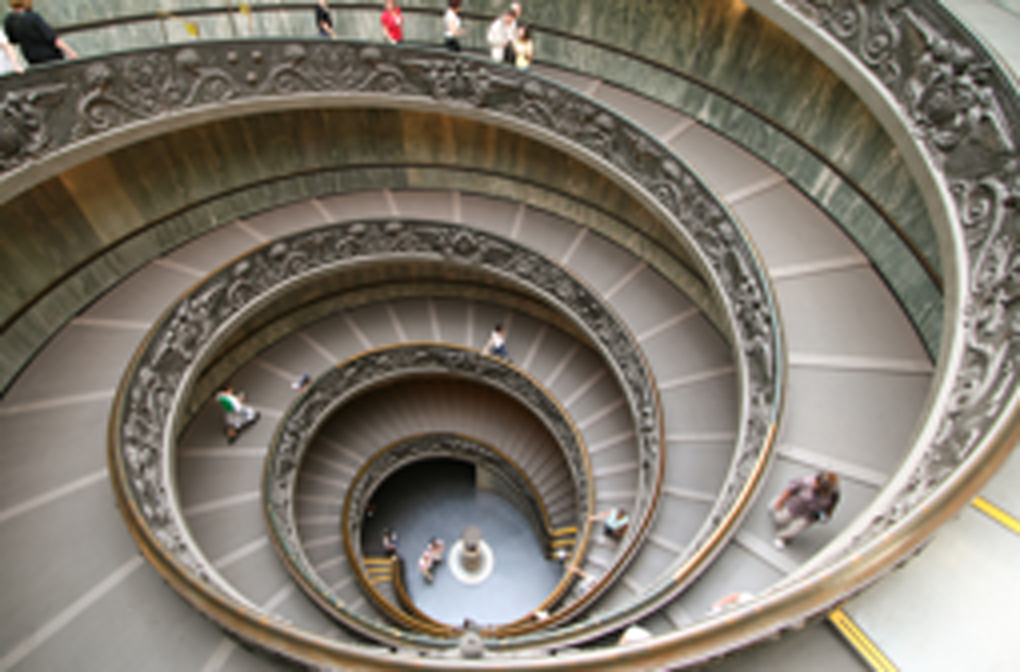}} 
& \scalebox{1.0}{\includegraphics[trim={11cm 15.0cm 11.5cm 0cm},clip]{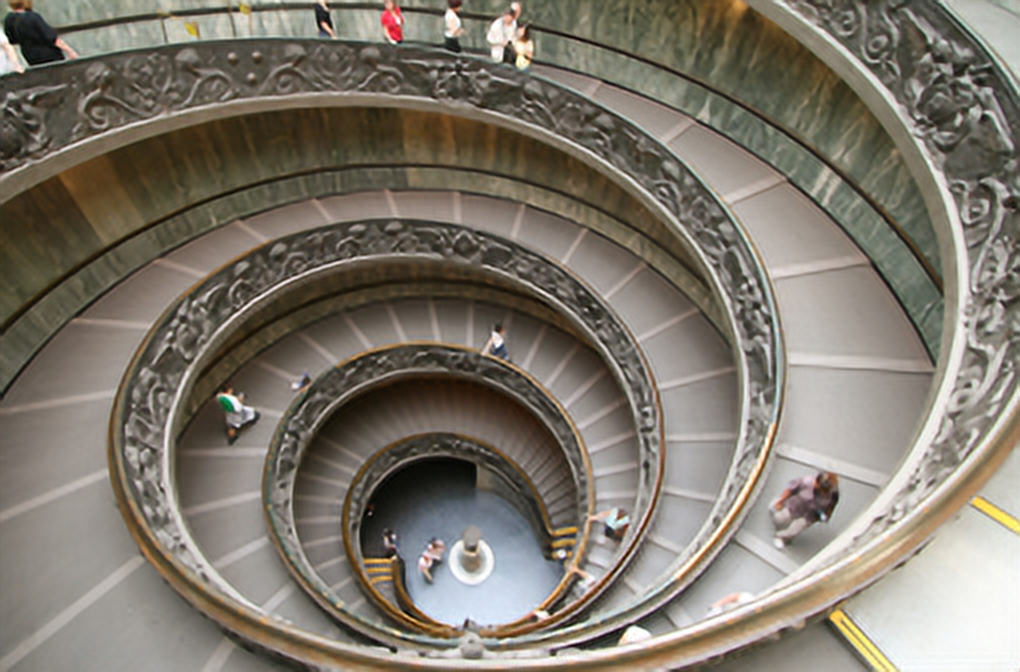}}
& \multicolumn{1}{|c|}{\scalebox{1.0}{\includegraphics[trim={11cm 15.0cm 11.5cm 0cm},clip]{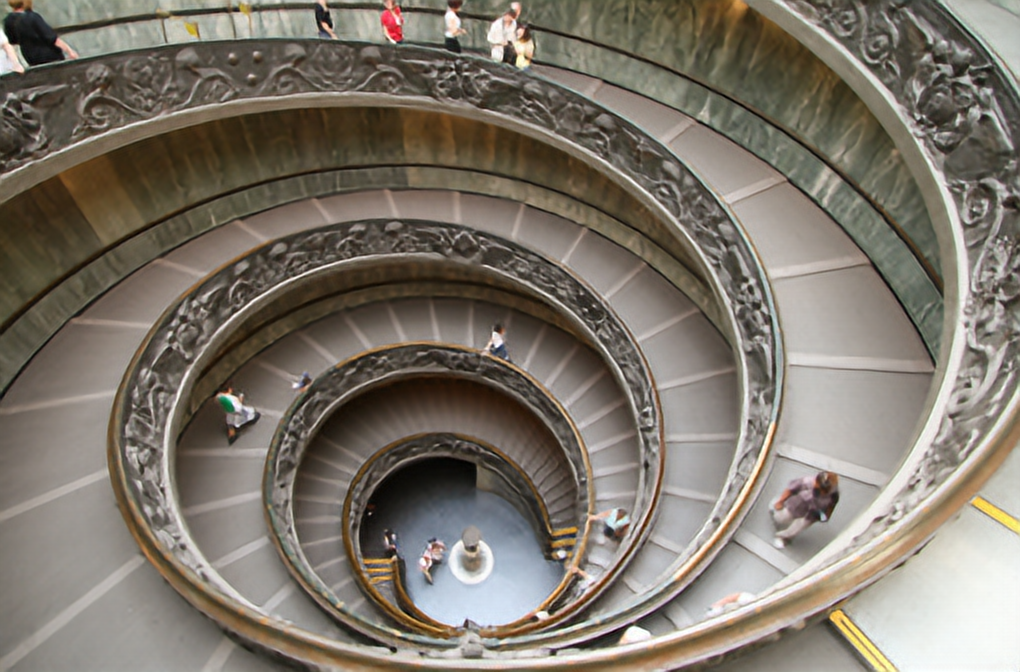}}}
\\
\multicolumn{2}{c}{}  
& \small{BICUBIC} & \small{MSE} & \multicolumn{1}{|c|}{\small{TLW+MSE}} \\

\multicolumn{2}{c}{\multirow{3}{*}[1.7cm]{\scalebox{.27}{\includegraphics{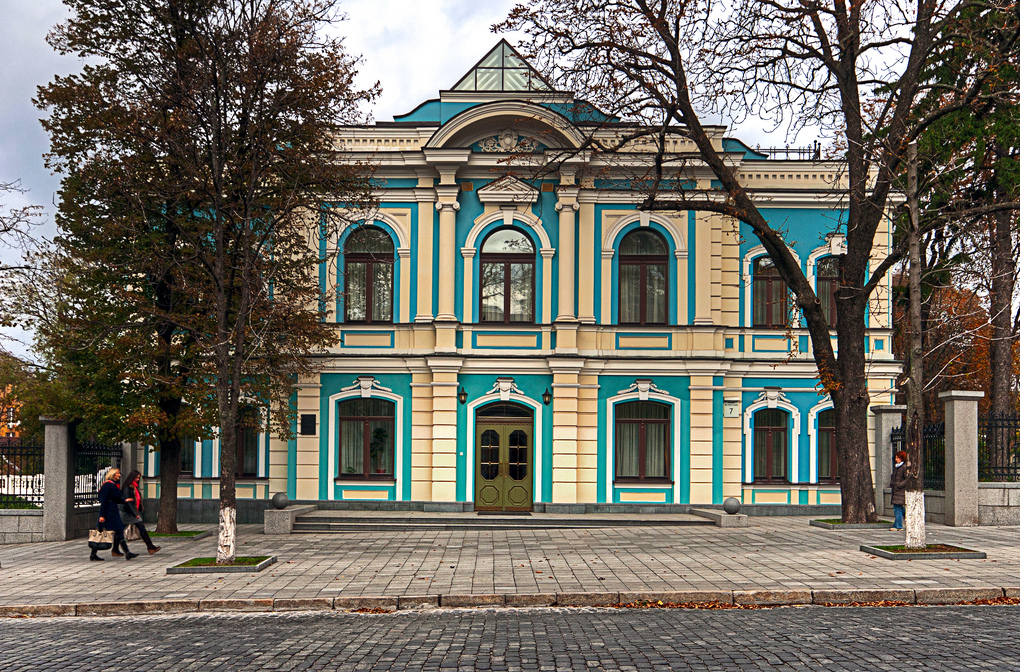}}}} 
& \scalebox{1.0}{\includegraphics[trim={11cm 5.0cm 11.5cm 10cm},clip]{pdf/18hr.png}} 
& \scalebox{1.0}{\includegraphics[trim={11cm 5.0cm 11.5cm 10cm},clip]{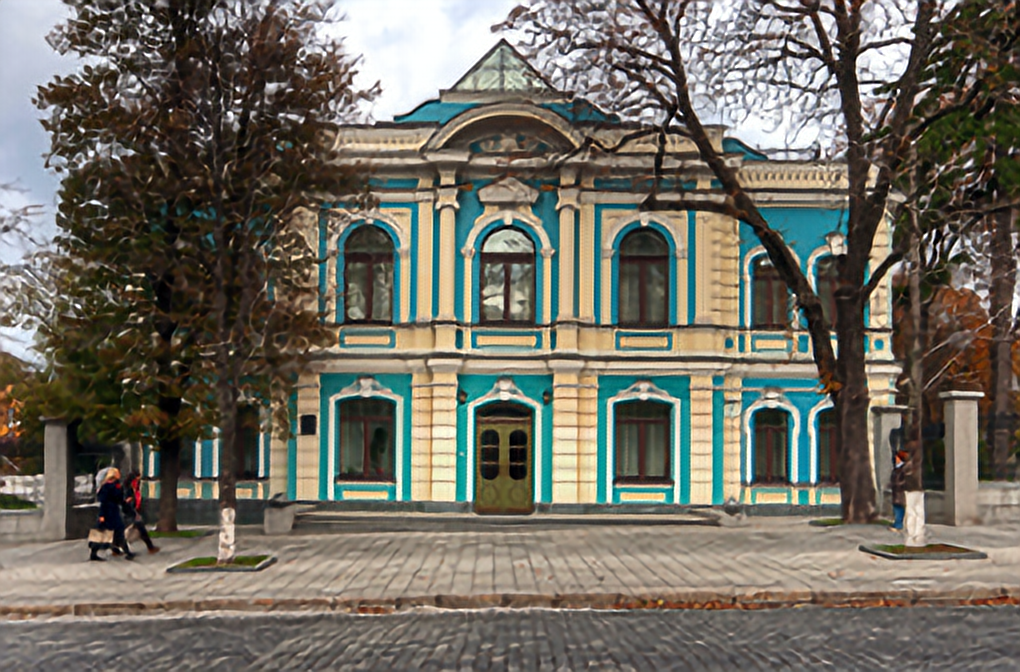}} 
& \multicolumn{1}{|c|}{\scalebox{1.0}{\includegraphics[trim={11cm 5.0cm 11.5cm 10cm},clip]{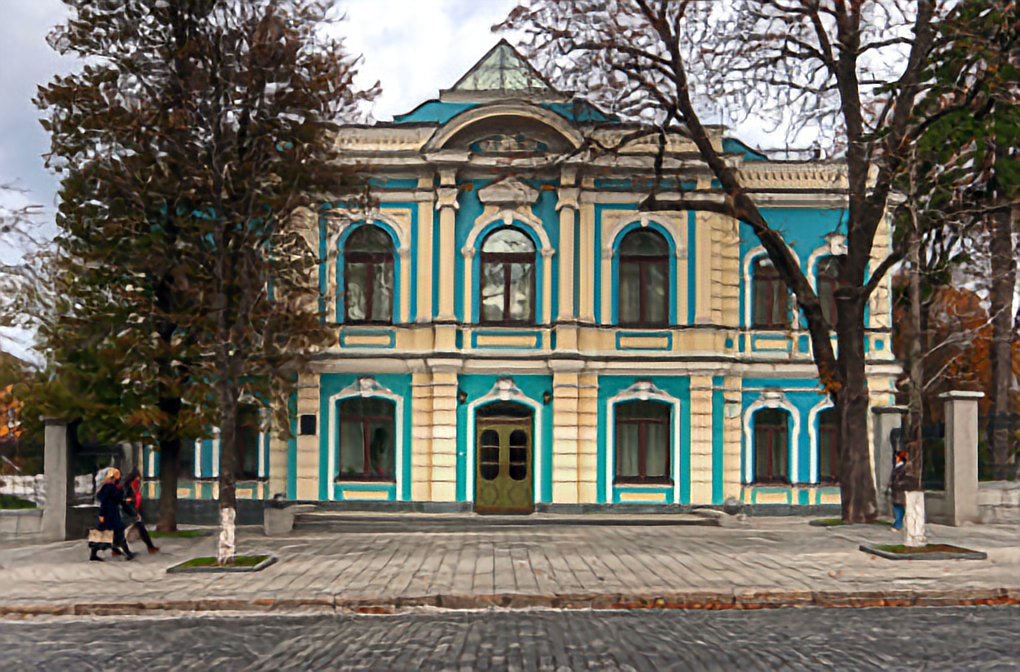}}}
\\
& 
& \small{HR} & \small{L1} & \multicolumn{1}{|c|}{\small{TLW+L1}} \\
&
& \scalebox{1.0}{\includegraphics[trim={11cm 5.0cm 11.5cm 10cm},clip]{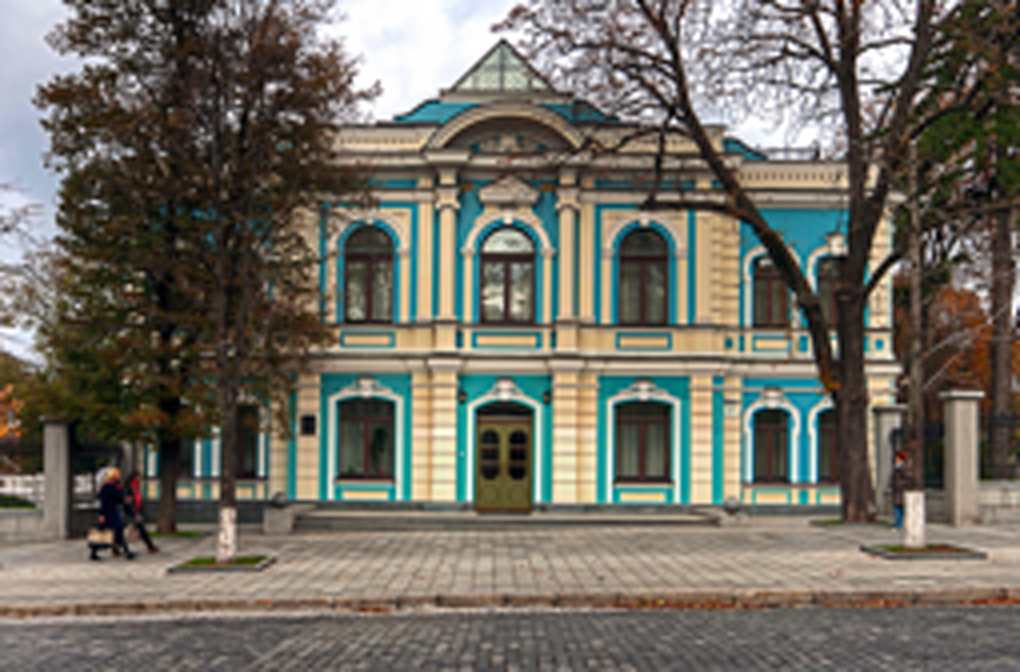}} 
& \scalebox{1.0}{\includegraphics[trim={11cm 5.0cm 11.5cm 10cm},clip]{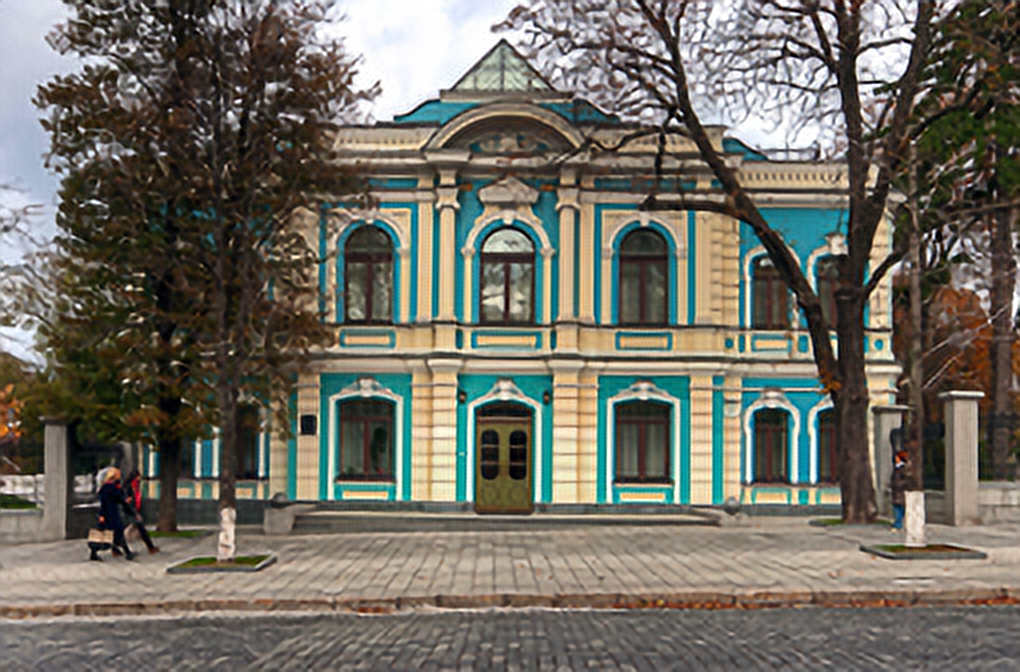}} 
& \multicolumn{1}{|c|}{\scalebox{1.0}{\includegraphics[trim={11cm 5.0cm 11.5cm 10cm},clip]{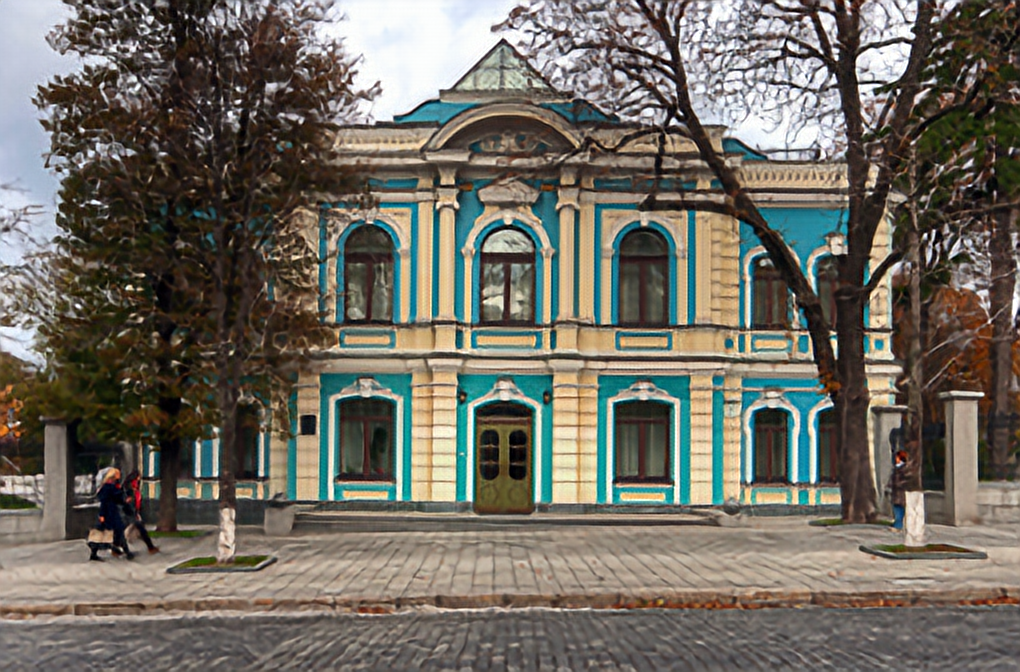}}}
\\
\multicolumn{2}{c}{} 
& \small{BICUBIC} & \small{MSE} & \multicolumn{1}{|c|}{\small{TLW+MSE}} \\
\cmidrule(lr){5-5}
    \end{tabular}
    \caption{Qualitative comparison of trained RCAN networks on weighted loss with the proposed method and unweighted loss as a result of super-resolution on x4 scale. }
    \label{fig:vis}
\end{figure*}

\begin{table*}[t]
\caption{Test results for unweighted L1, unweighted MSE, weighted (TLW) L1, and weighted (TLW) MSE. \textbf{Bold} indicates are the best performance between unweighted and weighted loss.}
\label{tbl:bench}
\centering
\begin{tabular}{ccc|cc|cc|cc|cc|cc}

\multirow{2}{*}{Scale}    & \multirow{2}{*}{Structure} & \multirow{2}{*}{Loss} & \multicolumn{2}{c}{Urban100}                           & \multicolumn{2}{c}{Set5}                               & \multicolumn{2}{c}{Set14}                              & \multicolumn{2}{c}{BSDS100}                            & \multicolumn{2}{c}{Manga109}                           \\ \cline{4-13} 
                     &                            &                       & \multicolumn{1}{c}{PSNR}             & LPIPS           & \multicolumn{1}{c}{PSNR}             & LPIPS           & \multicolumn{1}{c}{PSNR}             & LPIPS           & \multicolumn{1}{c}{PSNR}             & LPIPS           & \multicolumn{1}{c}{PSNR}             & LPIPS           \\ \hline
\multirow{13}{*}{x2} & \multirow{4}{*}{RCAN}      & L1+TLW                & \multicolumn{1}{c}{\textbf{30.3759}} & \textbf{0.1199} & \multicolumn{1}{c}{\textbf{36.6757}} & \textbf{0.0873} & \multicolumn{1}{c}{\textbf{32.2693}} & \textbf{0.1430} & \multicolumn{1}{c}{31.2668}          & \textbf{0.1542} & \multicolumn{1}{c}{\textbf{36.0380}} & \textbf{0.0523} \\ \cline{3-13} 
                     &                            & MSE+TLW               & \multicolumn{1}{c}{\textbf{30.3212}} & \textbf{0.1259} & \multicolumn{1}{c}{\textbf{37.0232}} & \textbf{0.0881} & \multicolumn{1}{c}{\textbf{32.5356}} & \textbf{0.1433} & \multicolumn{1}{c}{\textbf{31.7473}} & \textbf{0.1605} & \multicolumn{1}{c}{\textbf{36.6390}} & 0.0534          \\ \cline{3-13} 
                     &                            & L1                    & \multicolumn{1}{c}{30.3200}          & 0.1211          & \multicolumn{1}{c}{36.5458}          & 0.0891          & \multicolumn{1}{c}{32.2031}          & 0.1445          & \multicolumn{1}{c}{\textbf{31.3175}} & 0.1581          & \multicolumn{1}{c}{35.5993}          & 0.0540          \\ \cline{3-13} 
                     &                            & MSE                   & \multicolumn{1}{c}{30.2215}          & 0.1297          & \multicolumn{1}{c}{36.8614}          & 0.0896          & \multicolumn{1}{c}{32.5009}          & 0.1483          & \multicolumn{1}{c}{31.6905}          & 0.1650          & \multicolumn{1}{c}{36.1346}          & \textbf{0.0534} \\ \cline{2-13} 
                     & \multirow{4}{*}{VDSR}      & L1+TLW                & \multicolumn{1}{c}{30.4647}          & \textbf{0.1330} & \multicolumn{1}{c}{37.2049}          & \textbf{0.0951} & \multicolumn{1}{c}{32.7103}          & \textbf{0.1547} & \multicolumn{1}{c}{31.7538}          & \textbf{0.1712} & \multicolumn{1}{c}{36.7121}          & \textbf{0.0584} \\ \cline{3-13} 
                     &                            & MSE+TLW               & \multicolumn{1}{c}{\textbf{30.3858}} & \textbf{0.1318} & \multicolumn{1}{c}{\textbf{37.2355}} & \textbf{0.0916} & \multicolumn{1}{c}{\textbf{32.6706}} & \textbf{0.1506} & \multicolumn{1}{c}{\textbf{31.7569}} & \textbf{0.1661} & \multicolumn{1}{c}{\textbf{36.7320}} & \textbf{0.0558} \\ \cline{3-13} 
                     &                            & L1                    & \multicolumn{1}{c}{\textbf{30.4850}} & 0.1339          & \multicolumn{1}{c}{\textbf{37.2905}} & 0.0955          & \multicolumn{1}{c}{\textbf{32.7652}} & 0.1553          & \multicolumn{1}{c}{\textbf{31.7876}} & 0.1713          & \multicolumn{1}{c}{\textbf{36.8579}} & 0.0588          \\ \cline{3-13} 
                     &                            & MSE                   & \multicolumn{1}{c}{30.3689}          & 0.1332          & \multicolumn{1}{c}{37.1663}          & 0.0922          & \multicolumn{1}{c}{32.6692}          & 0.1518          & \multicolumn{1}{c}{31.7200}          & 0.1679          & \multicolumn{1}{c}{36.7230}          & 0.0568          \\ \cline{2-13} 
                     & \multirow{4}{*}{EDSR}      & L1+TLW                & \multicolumn{1}{c}{30.3618}          & \textbf{0.1259} & \multicolumn{1}{c}{\textbf{37.0381}} & \textbf{0.0903} & \multicolumn{1}{c}{32.4444}          & \textbf{0.1479} & \multicolumn{1}{c}{31.6571}          & \textbf{0.1625} & \multicolumn{1}{c}{\textbf{36.1764}} & \textbf{0.0537} \\ \cline{3-13} 
                     &                            & MSE+TLW               & \multicolumn{1}{c}{\textbf{30.1071}} & \textbf{0.1267} & \multicolumn{1}{c}{\textbf{36.9814}} & \textbf{0.0880} & \multicolumn{1}{c}{\textbf{32.3032}} & 0.1467          & \multicolumn{1}{c}{\textbf{31.5222}} & 0.1611          & \multicolumn{1}{c}{\textbf{35.8462}} & \textbf{0.0528} \\ \cline{3-13} 
                     &                            & L1                    & \multicolumn{1}{c}{\textbf{30.3724}} & 0.1269          & \multicolumn{1}{c}{36.9925}          & 0.0912          & \multicolumn{1}{c}{\textbf{32.5007}} & 0.1492          & \multicolumn{1}{c}{\textbf{31.6889}} & 0.1650          & \multicolumn{1}{c}{36.1681}          & 0.0545          \\ \cline{3-13} 
                     &                            & MSE                   & \multicolumn{1}{c}{30.0126}          & 0.1273          & \multicolumn{1}{c}{36.8242}          & 0.0880          & \multicolumn{1}{c}{32.2679}          & \textbf{0.1457} & \multicolumn{1}{c}{31.5106}          & \textbf{0.1604} & \multicolumn{1}{c}{35.6070}          & 0.0532          \\ \cline{2-13} 
                     & bicubic                    & -                     & \multicolumn{1}{c}{27.0711}          & 0.1934          & \multicolumn{1}{c}{33.9569}          & 0.1308          & \multicolumn{1}{c}{30.3250}          & 0.1912          & \multicolumn{1}{c}{29.7593}          & 0.2061          & \multicolumn{1}{c}{31.2655}          & 0.0923          \\ \hline
\multirow{13}{*}{x3} & \multirow{4}{*}{RCAN}      & L1+TLW                & \multicolumn{1}{c}{\textbf{26.7433}} & \textbf{0.2323} & \multicolumn{1}{c}{33.0242}          & \textbf{0.1587} & \multicolumn{1}{c}{29.1413}          & \textbf{0.2453} & \multicolumn{1}{c}{28.4921}          & \textbf{0.2700} & \multicolumn{1}{c}{\textbf{31.3582}} & \textbf{0.1188} \\ \cline{3-13} 
                     &                            & MSE+TLW               & \multicolumn{1}{c}{\textbf{26.6590}} & \textbf{0.2390} & \multicolumn{1}{c}{\textbf{33.0604}} & \textbf{0.1616} & \multicolumn{1}{c}{\textbf{29.2588}} & \textbf{0.2519} & \multicolumn{1}{c}{\textbf{28.6232}} & \textbf{0.2775} & \multicolumn{1}{c}{\textbf{31.3283}} & \textbf{0.1284} \\ \cline{3-13} 
                     &                            & L1                    & \multicolumn{1}{c}{26.7139}          & 0.2399          & \multicolumn{1}{c}{\textbf{33.1808}} & 0.1624          & \multicolumn{1}{c}{\textbf{29.2791}} & 0.2544          & \multicolumn{1}{c}{\textbf{28.6613}} & 0.2826          & \multicolumn{1}{c}{31.3425}          & 0.1226          \\ \cline{3-13} 
                     &                            & MSE                   & \multicolumn{1}{c}{26.4716}          & 0.2467          & \multicolumn{1}{c}{32.8152}          & 0.1665          & \multicolumn{1}{c}{29.1986}          & 0.2542          & \multicolumn{1}{c}{28.5739}          & 0.2789          & \multicolumn{1}{c}{31.0364}          & 0.1354          \\ \cline{2-13} 
                     & \multirow{4}{*}{VDSR}      & L1+TLW                & \multicolumn{1}{c}{\textbf{26.7056}} & \textbf{0.2542} & \multicolumn{1}{c}{\textbf{33.2478}} & \textbf{0.1730} & \multicolumn{1}{c}{\textbf{29.3746}} & \textbf{0.2650} & \multicolumn{1}{c}{\textbf{28.6638}} & \textbf{0.2943} & \multicolumn{1}{c}{\textbf{31.3448}} & \textbf{0.1350} \\ \cline{3-13} 
                     &                            & MSE+TLW               & \multicolumn{1}{c}{\textbf{26.5479}} & 0.2565          & \multicolumn{1}{c}{\textbf{33.0568}} & 0.1721          & \multicolumn{1}{c}{\textbf{29.2919}} & 0.2639          & \multicolumn{1}{c}{\textbf{28.5781}} & 0.2930          & \multicolumn{1}{c}{\textbf{31.2508}} & \textbf{0.1338} \\ \cline{3-13} 
                     &                            & L1                    & \multicolumn{1}{c}{26.5508}          & 0.2581          & \multicolumn{1}{c}{33.1737}          & 0.1741          & \multicolumn{1}{c}{29.3387}          & 0.2650          & \multicolumn{1}{c}{28.6052}          & 0.2952          & \multicolumn{1}{c}{31.2613}          & 0.1356          \\ \cline{3-13} 
                     &                            & MSE                   & \multicolumn{1}{c}{26.4821}          & \textbf{0.2548} & \multicolumn{1}{c}{32.9160}          & \textbf{0.1702} & \multicolumn{1}{c}{29.1317}          & \textbf{0.2616} & \multicolumn{1}{c}{28.5499}          & \textbf{0.2890} & \multicolumn{1}{c}{30.7235}          & 0.1355          \\ \cline{2-13} 
                     & \multirow{4}{*}{EDSR}      & L1+TLW                & \multicolumn{1}{c}{\textbf{26.8388}} & \textbf{0.2350} & \multicolumn{1}{c}{\textbf{33.3900}} & \textbf{0.1598} & \multicolumn{1}{c}{\textbf{29.4302}} & \textbf{0.2459} & \multicolumn{1}{c}{\textbf{28.7315}} & \textbf{0.2748} & \multicolumn{1}{c}{\textbf{31.6810}} & \textbf{0.1201} \\ \cline{3-13} 
                     &                            & MSE+TLW               & \multicolumn{1}{c}{\textbf{26.7724}} & \textbf{0.2343} & \multicolumn{1}{c}{\textbf{33.3424}} & 0.1564          & \multicolumn{1}{c}{\textbf{29.3464}} & \textbf{0.2469} & \multicolumn{1}{c}{\textbf{28.6931}} & 0.2711          & \multicolumn{1}{c}{\textbf{31.5225}} & \textbf{0.1178} \\ \cline{3-13} 
                     &                            & L1                    & \multicolumn{1}{c}{26.8047}          & 0.2370          & \multicolumn{1}{c}{33.2264}          & 0.1608          & \multicolumn{1}{c}{29.3188}          & 0.2505          & \multicolumn{1}{c}{28.6887}          & 0.2762          & \multicolumn{1}{c}{31.4931}          & 0.1206          \\ \cline{3-13} 
                     &                            & MSE                   & \multicolumn{1}{c}{26.5738}          & 0.2377          & \multicolumn{1}{c}{33.0845}          & \textbf{0.1548} & \multicolumn{1}{c}{29.1554}          & 0.2471          & \multicolumn{1}{c}{28.5710}          & \textbf{0.2689} & \multicolumn{1}{c}{31.0054}          & 0.1219          \\ \cline{2-13} 
                     & bicubic                    & -                     & \multicolumn{1}{c}{24.5817}          & 0.3054          & \multicolumn{1}{c}{30.6089}          & 0.2170          & \multicolumn{1}{c}{27.5664}          & 0.2951          & \multicolumn{1}{c}{27.3356}          & 0.3214          & \multicolumn{1}{c}{27.2258}          & 0.1821          \\ \hline
\multirow{13}{*}{x4} & \multirow{4}{*}{RCAN}      & L1+TLW                & \multicolumn{1}{c}{\textbf{24.8208}} & \textbf{0.3126} & \multicolumn{1}{c}{\textbf{30.6185}} & \textbf{0.2138} & \multicolumn{1}{c}{\textbf{27.3370}} & \textbf{0.3152} & \multicolumn{1}{c}{\textbf{26.8788}} & \textbf{0.3480} & \multicolumn{1}{c}{\textbf{28.3739}} & \textbf{0.1833} \\ \cline{3-13} 
                     &                            & MSE+TLW               & \multicolumn{1}{c}{\textbf{24.7003}} & \textbf{0.3194} & \multicolumn{1}{c}{30.2342}          & \textbf{0.2166} & \multicolumn{1}{c}{27.2306}          & \textbf{0.3205} & \multicolumn{1}{c}{26.8593}          & \textbf{0.3490} & \multicolumn{1}{c}{\textbf{28.0302}} & \textbf{0.2101} \\ \cline{3-13} 
                     &                            & L1                    & \multicolumn{1}{c}{24.6021}          & 0.3226          & \multicolumn{1}{c}{30.0164}          & 0.2206          & \multicolumn{1}{c}{27.1419}          & 0.3229          & \multicolumn{1}{c}{26.8588}          & 0.3584          & \multicolumn{1}{c}{27.7534}          & 0.1918          \\ \cline{3-13} 
                     &                            & MSE                   & \multicolumn{1}{c}{24.5678}          & 0.3368          & \multicolumn{1}{c}{\textbf{30.4734}} & 0.2366          & \multicolumn{1}{c}{\textbf{27.3460}} & 0.3282          & \multicolumn{1}{c}{\textbf{26.9782}} & 0.3598          & \multicolumn{1}{c}{27.8246}          & 0.2200          \\ \cline{2-13} 
                     & \multirow{4}{*}{VDSR}      & L1+TLW                & \multicolumn{1}{c}{24.6692}          & \textbf{0.3449} & \multicolumn{1}{c}{30.6993}          & \textbf{0.2321} & \multicolumn{1}{c}{27.3846}          & \textbf{0.3389} & \multicolumn{1}{c}{27.0428}          & \textbf{0.3790} & \multicolumn{1}{c}{28.0429}          & 0.2076          \\ \cline{3-13} 
                     &                            & MSE+TLW               & \multicolumn{1}{c}{\textbf{24.7450}} & \textbf{0.3436} & \multicolumn{1}{c}{\textbf{30.7785}} & \textbf{0.2324} & \multicolumn{1}{c}{\textbf{27.4842}} & \textbf{0.3390} & \multicolumn{1}{c}{\textbf{27.0925}} & \textbf{0.3779} & \multicolumn{1}{c}{\textbf{28.0901}} & \textbf{0.2075} \\ \cline{3-13} 
                     &                            & L1                    & \multicolumn{1}{c}{\textbf{24.6807}} & 0.3470          & \multicolumn{1}{c}{\textbf{30.7434}} & 0.2335          & \multicolumn{1}{c}{\textbf{27.4059}} & 0.3408          & \multicolumn{1}{c}{\textbf{27.0514}} & 0.3810          & \multicolumn{1}{c}{\textbf{28.1251}} & \textbf{0.2072} \\ \cline{3-13} 
                     &                            & MSE                   & \multicolumn{1}{c}{24.7006}          & 0.3438          & \multicolumn{1}{c}{30.6993}          & 0.2337          & \multicolumn{1}{c}{27.3946}          & 0.3393          & \multicolumn{1}{c}{27.0651}          & 0.3763          & \multicolumn{1}{c}{28.0091}          & 0.2081          \\ \cline{2-13} 
                     & \multirow{4}{*}{EDSR}      & L1+TLW                & \multicolumn{1}{c}{\textbf{24.8143}} & \textbf{0.3188} & \multicolumn{1}{c}{30.6716}          & 0.2174          & \multicolumn{1}{c}{\textbf{27.4417}} & \textbf{0.3189} & \multicolumn{1}{c}{\textbf{27.0938}} & \textbf{0.3526} & \multicolumn{1}{c}{\textbf{28.2490}} & \textbf{0.1901} \\ \cline{3-13} 
                     &                            & MSE+TLW               & \multicolumn{1}{c}{\textbf{24.7110}} & \textbf{0.3212} & \multicolumn{1}{c}{\textbf{30.5246}} & \textbf{0.2192} & \multicolumn{1}{c}{\textbf{27.3177}} & \textbf{0.3197} & \multicolumn{1}{c}{\textbf{26.9813}} & 0.3523          & \multicolumn{1}{c}{\textbf{28.0535}} & \textbf{0.1950} \\ \cline{3-13} 
                     &                            & L1                    & \multicolumn{1}{c}{24.8048}          & 0.3213          & \multicolumn{1}{c}{\textbf{30.6984}} & \textbf{0.2171} & \multicolumn{1}{c}{27.3974}          & 0.3209          & \multicolumn{1}{c}{27.0822}          & 0.3565          & \multicolumn{1}{c}{28.2025}          & 0.1908          \\ \cline{3-13} 
                     &                            & MSE                   & \multicolumn{1}{c}{24.6278}          & 0.3236          & \multicolumn{1}{c}{30.4268}          & 0.2218          & \multicolumn{1}{c}{27.2967}          & 0.3201          & \multicolumn{1}{c}{26.9653}          & \textbf{0.3477} & \multicolumn{1}{c}{27.8860}          & 0.2051          \\ \cline{2-13} 
                     & bicubic                    & -                     & \multicolumn{1}{c}{23.2383}          & 0.3838          & \multicolumn{1}{c}{28.5944}          & 0.2838          & \multicolumn{1}{c}{25.9904}          & 0.3677          & \multicolumn{1}{c}{26.0741}          & 0.4009          & \multicolumn{1}{c}{25.1164}          & 0.2547          \\ \hline

\end{tabular}
\end{table*}

\begin{table*}[t]
\caption[Comparing the trainable loss weights(TLW) with weighting loss based on uncertainty(UDL) ]{
The evaluation results of training the HAT\cite{HAT} network based on the proposed weighted loss ({L1-TLW}), weighted loss based on uncertainty ({L1-UDL}), and loss without weighting ({L1}) at a scale of {x4} on the {DIV2K} dataset are presented. The superior results are \textbf{highlighted} for each metric.}
\label{tbl:benchhat}
\centering

\begin{tabular}{ccc|cc|cc|cc|cc|cc}

{Scale} & {Structure} & {Loss} & \multicolumn{2}{c}{Urban100}                           & \multicolumn{2}{c}{Set5}                               & \multicolumn{2}{c}{Set14}  & \multicolumn{2}{c}{BSDS100}       & \multicolumn{2}{c}{Manga109} \\ \cline{4-13} 

      & &  & {PSNR}             & LPIPS           &{PSNR}             & LPIPS           & {PSNR}             & LPIPS           & {PSNR}             & LPIPS           & PSNR  & LPIPS       \\ \hline

\multirow{3}{*}{x4}  &  \multirow{3}{*}{HAT \protect\cite{HAT}}      & L1-TLW                & \textbf{26.1378} & \textbf{0.2768} & {32.2084} & \textbf{0.2053} & 28.3860 & {0.2986} & \textbf{27.6506} & \textbf{0.3422} & \textbf{30.7225} & \textbf{0.1570}\\   \cline{3-13} 

 & & L1-UDL\protect\cite{Uncertainty}             & {25.8823} & {0.2894} & {32.0091} & {0.2084} & 28.2914 & {0.3049} & 27.573 & {0.3480} & {30.3009} & {0.1603}\\   \cline{3-13} 

&  & L1                    & {26.0996} & {0.2800} & \textbf{32.2203} & {0.2059} & \textbf{28.4364} & \textbf{0.2981} & {27.6451} & {0.3440} & {30.6695} & {0.1579}\\ 
 
\hline

\end{tabular}
\end{table*}

\subsection{$k$ ratio during training process }
To investigate how the weighting network works and determine the ratio of the total weight to the size of the image, the changes in the output of the weighting network for the same input in different epochs have been investigated to clarify the features and performance of this network. As shown in Fig.\ref{fig:weight_epoch}, the network at the beginning of training, emphasizes the edges. In the 20th epoch, the weighting network tries to give more importance to the textures and to maintain the edge's importance. In the following epochs, the colors of some objects give higher weights and maintain the importance of edges and textures. 
The increase in the features that the weighting network has been emphasizing, indicates that the ratio of the total weight to the size of the image should increase so that the network can gain more flexibility to emphasize several new pixels during training. For this reason,  the ratio of the total weight is increased to the size of the image based on  $k = 0.6 \times (1 - e^{-epoch/200})+0.3$ equation, during the training time.

\subsection{Comparing trainable loss weights(TLW) with unweighted loss}
Three super-resolution models have been trained with weighted and unweighted loss, to compare the proposed and traditional loss. The RCAN \cite{RCAN}, VDSR \cite{VDSR}, and EDSR \cite{EDSR} models have been used to compare the proposed loss, similar to Ning et al \cite{Uncertainty} and Abrahamyan et al \cite{GVloss} studies. All models have been trained on the General100, BSDS200, and T91 datasets. Low-resolution images have been produced by the bicubic downsampler with 2, 3, and 4 scales. Models are evaluated on  Set5 , Set14 , Urban100, BSDS100, and Manga109 datasets. Also, The PSNR and LPIPS metrics are used to compare the models. In the evaluation state, PSNR was calculated based on the Y channel of the YCbCr image.
Each model has been trained in each scale with four different losses, in parallel. These losses include weighted L1, weighted MSE, L1, and MSE, respectively. The initial weight of these four models has been considered equal. In each iteration, four models are trained on the batch of HR-LR images. Also, the weighting model is trained based on the optimization method of Algorithm\ref{alg:one}, in each iteration.
As seen in Table\ref{tbl:bench}, the results on both proposed losses have been able to reach better results in most models, datasets, and scales for both metrics (LPIPS and PSNR). Also, as Fig.\ref{fig:vis} shows, the qualitative results of the proposed weighted loss are better and it has been able to better reconstruct the edges and textures of the HR image by giving different weights to different pixels of the image during training.
\subsection{Comparing the trainable loss weights(TLW) with weighted loss based on uncertainty(UDL)}
In another experience, a comparison was made with the study by \cite{Uncertainty}, which proposes uncertainty-based weights for training the super-resolution network. For this comparison, we used the HAT network architecture \cite{HAT}. The HAT network has achieved the highest PSNR compared to other deep neural network architectures by introducing a Hybrid Attention Transformer (HAT). To compare the proposed method, we trained three HAT networks: one with the proposed weighting loss(L1-TLW), one with the weighted loss proposed by \cite{Uncertainty} (L1-UDL), and one without weighting (L1). The results of these three networks were compared. According to \cite{Uncertainty}, to train the network with weighted loss based on uncertainty, it is necessary to add a three-layer network to the base network to estimate uncertainty. After sufficient training according to Eq. \ref{eq:unc_step1}, the network is trained to estimate uncertainty and the SR image simultaneously. Then, based on Eq. \ref{eq:unc_step2}, the network is trained using the estimated uncertainty-based weighting. 
To compare the three networks (TLW weighting, uncertainty-based weighting, no weighting), we trained all three networks on the DIV2K dataset. Then, all three networks were evaluated on the Set5, Set14, Urban100, BSDS100, and Manga109 datasets. The training images were divided into batches of size 366x366 and fed into the network. The LR images were generated using a bicubic downsampling algorithm with a scale factor of 4. The value of k was determined as shown in Figure 2. Each network was trained for 50 epochs on the entire training dataset. After the 10th epoch, uncertainty-based weighting was used instead of ESU loss for uncertaity-based weighting model. As the results in Table 1 indicate, our proposed method achieved better results compared to uncertainty-based weighting in almost all evaluation datasets. Additionally, the network obtained better results in terms of LPIPS and PSNR metrics compared to training without weighting in the majority of evaluation datasets.

\section{Conclusion}
In summary,  the effect of trainable loss weights (TLW) on image super-resolution has been investigated. A weight of loss criterion based on perceptual similarity has been proposed so that the different weights of loss can be compared. According to the results, it was indicated that the proposed weighting method is effective on various deep network architectures, datasets, and image degradation scales. Extensive experiments showed that using the TLW method to weight L1 and MSE losses restores the image more effectively than using unweighted losses and uncertainty-based weighted loss.



\begin{thebibliography}{1}
\bibliographystyle{IEEEtran}

\bibitem{realworldreview} Chen, Honggang, et al. "Real-world single image super-resolution: A brief review." Information Fusion 79 (2022): 124-145.

\bibitem{survey}Wang, Zhihao, Jian Chen, and Steven CH Hoi. "Deep learning for image super-resolution: A survey." IEEE transactions on pattern analysis and machine intelligence (2020).

\bibitem{LPIPS} Zhang, Richard, et al. "The unreasonable effectiveness of deep features as a perceptual metric." Proceedings of the IEEE conference on computer vision and pattern recognition. 2018.


\bibitem{EM} Dempster, Arthur P., Nan M. Laird, and Donald B. Rubin. "Maximum likelihood from incomplete data via the EM algorithm." Journal of the Royal Statistical Society: Series B (Methodological) 39.1 (1977): 1-22.

\bibitem{subpixel} Shi, Wenzhe, et al. "Real-time single image and video super-resolution using an efficient sub-pixel convolutional neural network." Proceedings of the IEEE conference on computer vision and pattern recognition. 2016.

\bibitem{metaSR} Hu, Xuecai, et al. "Meta-SR: A magnification-arbitrary network for super-resolution." Proceedings of the IEEE/CVF Conference on Computer Vision and Pattern Recognition. 2019.

\bibitem{VDSR}Kim, Jiwon, Jung Kwon Lee, and Kyoung Mu Lee. "Accurate image super-resolution using very deep convolutional networks." Proceedings of the IEEE conference on computer vision and pattern recognition. 2016.

\bibitem{RCAN}Zhang, Yulun, et al. "Image super-resolution using very deep residual channel attention networks." Proceedings of the European conference on computer vision (ECCV). 2018.

\bibitem{HAT} Chen, X., et al. "Activating More Pixels in Image Super-Resolution Transformer." arXiv preprint arXiv:2205.04437. 2022.

\bibitem{unfolding} Zhang, Kai, Luc Van Gool, and Radu Timofte. "Deep unfolding network for image super-resolution." Proceedings of the IEEE/CVF Conference on Computer Vision and Pattern Recognition. 2020.

\bibitem{denoiserprior} Zhang, Kai, et al. "Learning deep CNN denoiser prior for image restoration." Proceedings of the IEEE conference on computer vision and pattern recognition. 2017.

\bibitem{multipledegrad} Zhang, Kai, Wangmeng Zuo, and Lei Zhang. "Learning a single convolutional super-resolution network for multiple degradations." Proceedings of the IEEE Conference on Computer Vision and Pattern Recognition. 2018.

\bibitem{kernelGAN}Bell-Kligler, Sefi, Assaf Shocher, and Michal Irani. "Blind super-resolution kernel estimation using an internal-gan." arXiv preprint arXiv:1909.06581 (2019).

\bibitem{IKC} Gu, Jinjin, et al. "Blind super-resolution with iterative kernel correction." Proceedings of the IEEE/CVF Conference on Computer Vision and Pattern Recognition. 2019.

\bibitem{SRGAN} Ledig, Christian, et al. "Photo-realistic single image super-resolution using a generative adversarial network." Proceedings of the IEEE conference on computer vision and pattern recognition. 2017.

\bibitem{cincGAN} Yuan, Yuan, et al. "Unsupervised image super-resolution using cycle-in-cycle generative adversarial networks." Proceedings of the IEEE Conference on Computer Vision and Pattern Recognition Workshops. 2018.

\bibitem{Esrgan} Wang, Xintao, et al. "Esrgan: Enhanced super-resolution generative adversarial networks." Proceedings of the European conference on computer vision (ECCV) workshops. 2018.

\bibitem{GRAM} Lee, Changwoo, and Ki-Seok Chung. "GRAM: Gradient rescaling attention model for data uncertainty estimation in single image super resolution." 2019 18th IEEE International Conference On Machine Learning And Applications (ICMLA). IEEE, 2019.

\bibitem{Investigat} Jo, Younghyun, Sejong Yang, and Seon Joo Kim. "Investigating loss functions for extreme super-resolution." Proceedings of the IEEE/CVF Conference on Computer Vision and Pattern Recognition Workshops. 2020.

\bibitem{Uncertainty} Ning, Qian, et al. "Uncertainty-Driven Loss for Single Image Super-Resolution." Advances in Neural Information Processing Systems 34 (2021).

\bibitem{Srobb} Rad, Mohammad Saeed, et al. "Srobb: Targeted perceptual loss for single image super-resolution." Proceedings of the IEEE/CVF International Conference on Computer Vision. 2019.

\bibitem{histogram} Mustafa, Aamir, Hongjie You, and Rafal K. Mantiuk. "A Comparative Study on the Loss Functions for Image Enhancement Networks." London Imaging Meeting. Vol. 3. No. 1. IS\&T 7003 Kilworth Lane, Springfield, VA 22151 USA: Society for Imaging Science and Technology, 2022.

\bibitem{GVloss} Abrahamyan, Lusine, et al. "Gradient variance loss for structure-enhanced image super-resolution." ICASSP 2022-2022 IEEE International Conference on Acoustics, Speech and Signal Processing (ICASSP). IEEE, 2022.

\bibitem{EDSR} Lim, Bee, et al. "Enhanced deep residual networks for single image super-resolution." Proceedings of the IEEE conference on computer vision and pattern recognition workshops. 2017.

\bibitem{MVB} Wang, Xi, and Junming Yin. "Relaxed multivariate Bernoulli distribution and its applications to deep generative models." Conference on Uncertainty in Artificial Intelligence. PMLR, 2020.

\end{thebibliography}
\end{document}